\documentclass[journal]{IEEEtran}

\usepackage{cite}
\usepackage{amsmath,amssymb,amsfonts}
\usepackage{algorithmic}
\usepackage{graphicx}
\usepackage{textcomp}
\usepackage{balance}
\usepackage{blindtext}
\usepackage{array}
\usepackage{ragged2e}
\usepackage{tabularx}

\begin{document}
\title{Towards open and expandable cognitive AI architectures for large-scale multi-agent human-robot collaborative learning}
\author{Georgios~Th.~Papadopoulos$^{1}$,~\IEEEmembership{Member,~IEEE,}~Margherita~Antona$^{1}$~\IEEEmembership{}and~Constantine~Stephanidis$^{12}$~\IEEEmembership{}\\
$^{1}$Institute of Computer Science (ICS), Foundation for Research and Technology - Hellas (FORTH), Greece\\
$^{2}$Computer Science Department (CSD), University of Crete (UoC), Greece\\
Email: $\{$gepapado,antona,cs$\}$@ics.forth.gr}
\maketitle

\begin{abstract}
Learning from Demonstration (LfD) constitutes one of the most robust methodologies for constructing efficient cognitive robotic systems. Despite the large body of research works already reported, current key technological challenges include those of multi-agent learning and long-term autonomy. Towards this direction, a novel cognitive architecture for multi-agent LfD robotic learning is introduced in this paper, targeting to enable the reliable deployment of open, scalable and expandable robotic systems in large-scale and complex environments. In particular, the designed architecture capitalizes on the recent advances in the Artificial Intelligence (AI) (and especially the Deep Learning (DL)) field, by establishing a Federated Learning (FL)-based framework for incarnating a multi-human multi-robot collaborative learning environment. The fundamental conceptualization relies on employing multiple AI-empowered cognitive processes (implementing various robotic tasks) that operate at the edge nodes of a network of robotic platforms, while global AI models (underpinning the aforementioned robotic tasks) are collectively created and shared among the network, by elegantly combining information from a large number of human-robot interaction instances. Regarding pivotal novelties, the designed cognitive architecture a) introduces a new FL-based formalism that extends the conventional LfD learning paradigm to support large-scale multi-agent operational settings, b) elaborates previous FL-based self-learning robotic schemes so as to incorporate the human in the learning loop and c) consolidates the fundamental principles of FL with additional sophisticated AI-enabled learning methodologies for modelling the multi-level inter-dependencies among the robotic tasks. The applicability of the proposed framework is explained using an example of a real-world industrial case study (subject to ongoing research activities) for agile production-based Critical Raw Materials (CRM) recovery.
\end{abstract}


\section{Introduction}
\label{sec:Introduction}
Robotic learning through direct interaction with humans constitutes a well-established and highly active research field over the past few decades. In particular, the most dominant learning paradigm in this area, the so-called {\it{Learning from Demonstration}} (LfD), relies on the fundamental principle of robots acquiring new skills by learning to imitate a (human) teacher \cite{Ravichandar20}\cite{Argall09}. LfD (equally termed programming by demonstration, imitation learning, behavioral cloning or apprenticeship learning) concerns multiple aspects of robotics technology, including human-robot interaction, machine learning, machine vision and motor control \cite{furuta2020motion}\cite{seleem2020development}. Key advantageous characteristics of LfD that have contributed to its widespread adoption and successful application to diverse domains include \cite{Billard16}: 
\begin{itemize}
\item It enables robot programming by non-expert users;
\item It allows time-efficient learning, where task requirements are implicitly learned through demonstrations (and not by explicitly specifying all sequences of robotic actions);
\item It enables adaptive robotic behaviors;
\item It renders feasible for the robots to operate in complex, unstructured and time-varying environments.
\end{itemize}

So far, multiple and fundamentally different methods have been investigated and materialized in robotic platforms for capturing, modelling and learning from human feedback in numerous robotic manipulation tasks of varying complexity \cite{Zhu18}. Depending on the employed human demonstration means, LfD approaches generally fall into three main categories \cite{Ravichandar20}: 
\begin{itemize}
\item Kinesthetic teaching, which enables the human user to demonstrate by physically moving the robot through the desired motions \cite{Caccavale19}\cite{Capurso17};
\item Teleoperation, which requires an external input to guide the robot through a joystick, graphical user interface or other means \cite{Zhang18}\cite{Xu18};
\item Passive observation, where the robot learns from passively observing the user behavior \cite{Liu18}\cite{Finn17}.
\end{itemize}

A critical aspect in the design and deployment of any LfD scheme concerns the adopted methodology for refining the robot learned policies. In particular, different types of approaches have been introduced \cite{Hussein17}:
\begin{itemize}
\item Reinforcement learning, where a policy to solve a problem is learned via trial and error \cite{Zhang16}\cite{Silver16};
\item Optimization, which targets to find an optimal solution based on given criteria \cite{Cheng15} \cite{Hussieny15};
\item Transfer learning, where knowledge of a task or a domain is used to enhance the learning procedure for another task \cite{Brys15};
\item Apprenticeship learning, where demonstrated samples are used as a template for the desired performance \cite{Wulfmeier15};
\item Active learning, where the robotic agent is able to query an expert for the optimal response to a given state and to use these active samples to improve its policy \cite{Judah12};
\item Structured predictions, which is based on the fundamental consideration that an action is regarded as a sequence of dependent predictions \cite{Droniou14}.
\end{itemize}

Regarding implementation and deployment aspects of the designed LfD mechanisms, these have been dominated by the adoption of Machine Learning (ML) techniques, whose continuously increasing modelling and representation learning capabilities have correspondingly led to more robust and fine-grained LfD learning potentials. In particular, multiple and diverse ML approaches have been employed, ranging from Gaussian Mixture Models (GMMs) \cite{Wang16}, Hidden Markov Models (HMMs) \cite{Rozo13} and Dynamic Movement Primitives (DMPs) \cite{Li15} to, more recently introduced, Recurrent Neural Networks (RNNs) \cite{Rahmatizadeh18} and Convolutional Neural Networks (CNNs) \cite{Devin18}, to name a few. 

Despite the plethora of works in the LfD field, critical research challenges of paramount importance, namely {\bf{multi-agent learning}} \cite{Hussein17} and {\bf{long-term autonomy}} \cite{Kunze18}, need to be reliably addressed, in order to promote the wide-spread use of robots in open and complex environments. Regarding the former, robust solutions to multi-agent learning would enable, among others: a) sharing and aggregation of diverse knowledge/experiences among a large number of agents (especially in a multi-robot multi-human setting), b) creation of large volumes of heterogeneous and diversified datasets for training purposes, c) convergence to the development of more reliable and generalizable modules for robotic task execution and d) more efficient handling of complex tasks. On the other hand, enhancing long-term autonomy (i.e. enabling the robot to remain operational for as long as possible) would allow, among others: a) handling (short-, medium- or long-term) changes in the operational environment, b) managing parts of the environment that may not be fully known before system deployment or when new elements appear, c) addressing changes in user requirements (e.g. alternation of targeted tasks or how the robot should accomplish them) and d) adaptation to new knowledge (e.g. about the environment, provided by the human user, etc.) as it becomes available. The above need to be also investigated in conjunction with typical challenges in Human–Robot Interaction (HRI) schemes, like developing appropriate user interfaces, variance in human performance, variability in knowledge across human subjects, learning from noisy/imprecise human input, learning from very large or very sparse datasets, incremental learning, etc.

In this paper, a novel cognitive architecture for large-scale LfD robotic learning is introduced, targeting to reliably address the currently most critical challenges (and of outstanding importance) in the field, namely those of multi-agent learning, long-term autonomy and deployment of open, scalable and expandable robotic systems. The designed architecture takes advantage of the recent advances in the Artificial Intelligence (AI) (and especially the Deep Learning (DL)) field, by establishing a Federated Learning (FL)-based framework for incarnating a multi-human multi-robot collaborative learning environment. The fundamental conceptualization relies on employing multiple AI-empowered cognitive processes \cite{papadopoulos2017CVPR}\cite{hu2019irobot}\cite{wan2020cognitive}\cite{papadopoulos2016human} (implementing various robotic tasks, like sensing, navigation, manipulation, control, human-robot interaction, etc.) that operate at the edge nodes of a network of robotic platforms (i.e. adopting a decentralized edge computing setting), while global AI models (underpinning the aforementioned robotic tasks) are collectively created and shared among the network, by elegantly combining information from a large number of human-robot interaction instances. The designed cognitive AI architecture exhibits the following main novel advantageous characteristics, which significantly broaden the current capabilities of LfD learning schemes:
\begin{itemize}
\item It extends the current conventional LfD robotic learning paradigm to support large-scale multi-agent operational settings, by introducing a new FL-based formalism, which appropriately amplifies the algorithmic aspects of the conventional FL mechanism;
\item It elaborates previous FL-based self-learning robotic schemes, by incorporating the human in the learning loop (using intuitive and informative HRI mechanisms), while including diverse strategies for fine-grained adaptive analysis and integration of multi-human feedback in the Neural Network (NN) parameter update process, namely user weighting, parameter weighting and user clustering;
\item It consolidates the fundamental principles of FL with additional sophisticated AI-enabled learning methodologies for modelling the inter-dependencies among the robotic tasks and, hence, further reinforcing the robot knowledge$/$skill acquisition capabilities, namely transfer-, multitask- and meta-learning techniques.
\end{itemize}
Overall, the designed cognitive AI architecture essentially introduces a large-scale comprehensive human-robot collective intelligence scheme. An example of a real-world industrial case study (subject to ongoing research activities) for agile production-based Critical Raw Materials (CRM) recovery is investigated for explaining the applicability of the proposed framework. 

The remainder of the paper is organized as follows: Section \ref{sec:ChallengesLfD} details challenges and open issues currently present in the LfD learning field. Section \ref{sec:FL} briefly outlines the fundamental principles of the federated learning paradigm, while Section \ref{sec:FLRobotics} discusses prior FL-based works in robotics. Additionally, Section \ref{sec:CognitiveArchitecture} details the introduced AI-empowered multi-agent LfD cognitive architecture, whose applicability is demonstrated using a real-world industrial CRM recovery case study in Section \ref{sec:CRMrecovery}. Finally, conclusions and proposed future research directions, according to the designed cognitive architecture, are discussed in Section \ref{sec:Conclusions}.

\section{Challenges in LfD learning}
\label{sec:ChallengesLfD}

As outlined in Section \ref{sec:Introduction}, LfD constitutes a dominant methodology for robotic learning, due to the multiple advantageous characteristics that it exhibits \cite{Argall09}, such as that it allows non-expert robot programming, it typically requires a relatively small number of expert demonstrations, it allows controlled and safe learning circumstances for humans, it provides strong convergence guarantees, it is robotic platform independent, etc. Regarding the operational scenarios where LfD has so far shown significant advances, these involve the use of both manipulator (e.g. manufacturing \cite{vogt2017system}, assistive \cite{bhattacharjee2019towards}, healthcare \cite{wang2016motion}, social \cite{hayes2016autonomously}, etc.) and mobile (e.g. ground \cite{pan2017agile}, aerial \cite{loquercio2018dronet}, Bi/Quadru-pedal \cite{calandra2014bayesian}, underwater \cite{somers2016human}, etc.) robots. With respect to the overall LfD process outcome, this can be categorized to different levels of abstraction \cite{Ravichandar20}, including the learning of:
\begin{itemize}
\item Policies, i.e. the estimation of a function that generates the desired behavior \cite{ho2016generative};
\item Cost and reward functions, where it is assumed that the ideal behavior results from the optimization of a hidden function \cite{ravichandar2019skill};
\item Plans, i.e. high level structured schemes, composed of several sub-tasks or primitive actions \cite{mohseni2019simultaneous};
\item Multiple outcomes simultaneously, by jointly modeling complex behaviors at multiple levels of abstraction \cite{krishnan2019swirl}.
\end{itemize}

Despite the extensive research efforts devoted and the large body of corresponding outcomes produced, key technological challenges and open issues still remain to be robustly addressed, including, among others, the following ones \cite{Ravichandar20}\cite{Hussein17}\cite{Billard16}\cite{Zhu18}: 
\begin{itemize}
\item To involve in a more robust way a broad number of teachers with different styles of and possibly conflicting demonstrations, including the case of teachers with diverse idiosyncrasies and varying levels of expertise/experience; 
\item To transfer skills across multiple agents, including multiple and diverse types of robots, since so far research has been reduced to transfer of navigation or communication skills across swarms of relatively simple mobile robots;
\item To simultaneously learn multiple complex tasks, while storing and reusing prior knowledge at large scale, i.e. moving away from the typical scenario of focusing on a single task (or a set of closely related tasks) and a tabula rasa setting;
\item To robustly implement incremental learning schemes, i.e.  to enable the robot to select information, to reduce redundant data, to select features/representations and to store new data efficiently;
\item To reinforce the generalization ability, i.e. to enable the robots to learn from and to respond to stimuli unseen, yet similar, during training;
\item To model compound tasks, i.e. to jointly learn high-level task plans and low-level primitives, while also making use of multi-modal demonstrations;
\item To implement multi-agent imitation learning schemes, i.e. enabling each robotic agent to acquire knowledge and skills from a number of human teachers or other agents in a shared environment;
\item To operate in realistic, dynamic, time-varying and complex environments, aiming to enhance the long-term autonomous functioning of robots.
\end{itemize}

The recent outstanding advances in the AI field, especially concerning the collaborative FL paradigm and associated AI-empowered advanced learning methodologies, can provide fertile ground and highly promising research directions towards serving as enablers for providing robust solutions to the above-mentioned LfD robotic learning challenges. Especially with respect to the continuously growing need for deploying robotic solutions in large-scale, dynamic and complex environments, the demonstrated increased AI capabilities can provide reliable means for encountering the particularly crucial aspects of multi-agent learning \cite{Hussein17} (i.e. enabling robotic agents to acquire skills from their interaction with multiple human demonstrators as well as to exchange knowledge with other robots in a shared environment) and long-term autonomy \cite{Kunze18} (i.e. enabling robotic systems to perform autonomously in real-world scenarios over extended time periods).

\section{Federated learning paradigm}
\label{sec:FL}

Federated learning (or collaborative learning) is a ML paradigm where multiple computational nodes (e.g. computer clusters, PCs, mobile devices, etc.) collaboratively train a global (AI) model under the supervision of and orchestration by a central process (e.g. server, service provider, High-Performance Computing (HPC) infrastructure, etc.), without exchanging data among the nodes of the network (i.e. maintaining the training data of each node locally, in a decentralized way) \cite{Kairouz19}\cite{aledhari2020federated}\cite{ye2020edgefed}. Especially the latter characteristic renders FL a by-definition privacy-aware method, which can reliably mitigate many of the systemic privacy risks and costs resulting from traditional centralized ML \cite{Geyer17}\cite{lu2020privacy}\cite{zhu2020privacy}.

\begin{figure}[t]
\centering
\includegraphics[width=0.45\textwidth]{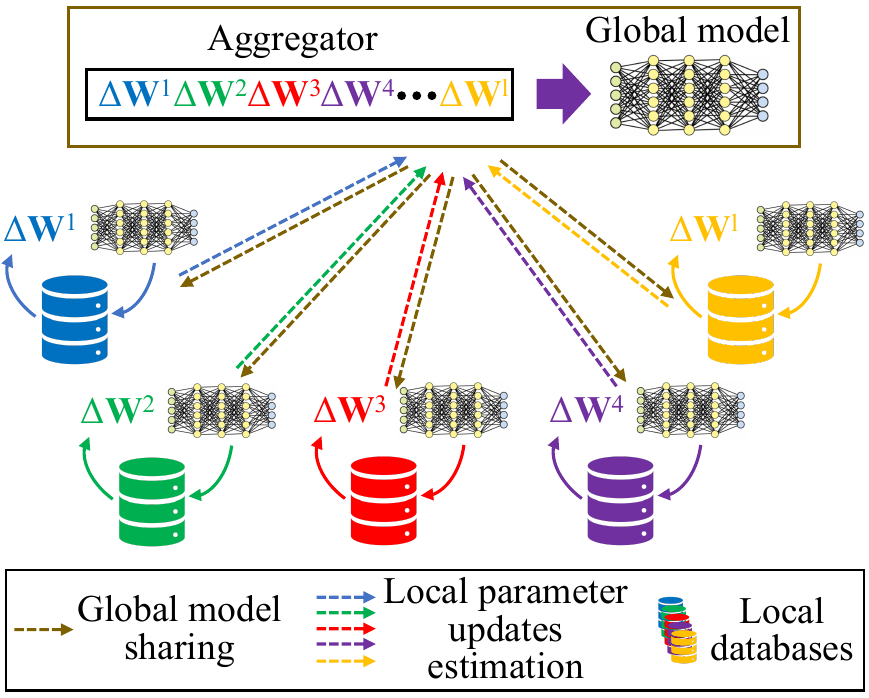}
\caption{Fundamental mechanism of the federated learning paradigm.}
\label{f:FL}
\end{figure}

The fundamental mechanism of the FL paradigm is illustrated in Fig. \ref{f:FL}. In particular, a global AI model, which is aimed to be collaboratively trained and shared among the network, is initially constructed using proxy data (either offline or at the central node). Then, the model is made available to the network and downloaded by each node. Every node encapsulates a local database that is used to estimate improved updates of the global model parameters (e.g. using conventional Stochastic Gradient Descent (SGD) for the case of NN-based AI modules), without making the local (federated) data available to the network. The computed local parameter updates (denoted $\Delta \textbf{W}^l$ in Fig. \ref{f:FL}) are asynchronously transmitted back to the central node (using encrypted communication), where an aggregation mechanism is responsible for combining them (often adopting a simple averaging operator) and periodically producing a new version of the global model. The overall process is iterative (i.e. continuously estimating improved versions of the global model) and can involve a very large number of heterogeneous computational nodes \cite{Bonawitz19}. It needs to be highlighted that each local computational node can maintain a customized version of the global model, while using the locally stored data for fine-tuning purposes.

FL-based training in highly heterogeneous and massive networks results into new technological challenges, related to the fields of large-scale ML, distributed optimization and privacy-preserving data analysis, which are summarized as follows \cite{Li20}:
\begin{itemize}
\item Communication efficiency, where FL-based systems need to reliably address the potentially massive number of available nodes and the corresponding slow network communication (by many orders of magnitude) compared to local computational speed;
\item Systems heterogeneity, where particular attention needs to be given towards balancing the varying storage, computational and communication capabilities of each node, due to the corresponding variability in hardware, network connectivity and power resources;
\item Statistical heterogeneity, which refers to the adopted data-generation paradigm in FL systems that typically violates frequently used independent and identically distributed (i.i.d.) assumptions in distributed optimization, which adds to the system design problem complexity;
\item Privacy, where an optimal trade-off needs to be made between FL-system efficiency and protecting sensitive information (throughout the training process) that can be revealed to a third-party or the central node.
\end{itemize}

\section{Federated learning in robotics}
\label{sec:FLRobotics}

\begin{table*}[t]
\caption{Comparison of architectural characteristics of multi-agent FL-based robotic systems of the literature}
\setlength\tabcolsep{1pt}
\begin{center}
\begin{tabular}{|c||c|c|c|c|c|c|c|}
\hline
&\multicolumn{7}{c|}{\begin{tabular}{c}Characteristics\end{tabular}}\\
\cline{2-8}
\centering\begin{tabular}{c}Robotic\\architecture\end{tabular}&
\begin{tabular}{c}Robot\\types\end{tabular}&
\begin{tabular}{c}Sensor\\types\end{tabular}&
\begin{tabular}{c}Task\\types\end{tabular}&
\begin{tabular}{c}Learning\\methodology\end{tabular}&
\begin{tabular}{c}Integration\\of human\end{tabular}&
\begin{tabular}{c}Advanced\\learning\\methodology\end{tabular}&
\begin{tabular}{c}Implemented\end{tabular}\\
\hline
\hline
\begin{tabular}{c}\cite{Liu19lifelong}\end{tabular}&
\begin{tabular}{c}Single\\(mobile\\robots)\end{tabular}&
\begin{tabular}{c}Single\\(visual)\end{tabular}&
\begin{tabular}{c}Single\\(navigation)\end{tabular}&
\begin{tabular}{c}Reinforcement\\learning\end{tabular}&
\begin{tabular}{c}No\end{tabular}&
\begin{tabular}{c}\textbf{Lifelong}\\\textbf{learning}\end{tabular}&
\begin{tabular}{c}\textbf{Yes}\\(real-world,\\simulation)\end{tabular}\\
\hline
\begin{tabular}{c}\cite{liang2019federated}\end{tabular}&
\begin{tabular}{c}Single\\(autonomous\\cars)\end{tabular}&
\begin{tabular}{c}Single\\(lidar)\end{tabular}&
\begin{tabular}{c}Single\\(Autonomous\\driving)\end{tabular}&
\begin{tabular}{c}Reinforcement\\learning\end{tabular}&
\begin{tabular}{c}No\end{tabular}&
\begin{tabular}{c}-\end{tabular}&
\begin{tabular}{c}\textbf{Yes}\\(real-world,\\simulation)\end{tabular}\\
\hline
\begin{tabular}{c}\cite{liu2020federated}\end{tabular}&
\begin{tabular}{c}Single\\(autonomous\\cars)\end{tabular}&
\begin{tabular}{c}\textbf{Multiple}\\(visual)\end{tabular}&
\begin{tabular}{c}Single\\(self-\\driving)\end{tabular}&
\begin{tabular}{c}Conventional\end{tabular}&
\begin{tabular}{c}No\end{tabular}&
\begin{tabular}{c}\textbf{Transfer}\\\textbf{learning}\end{tabular}&
\begin{tabular}{c}\textbf{Yes}\\(simulation)\end{tabular}\\
\hline
\begin{tabular}{c}\cite{khamidehi2020federated}\end{tabular}&
\begin{tabular}{c}Single\\(UAVs)\end{tabular}&
\begin{tabular}{c}Single\\(network\\connectivity)\end{tabular}&
\begin{tabular}{c}Single\\(path\\planning)\end{tabular}&
\begin{tabular}{c}Conventional\end{tabular}&
\begin{tabular}{c}No\end{tabular}&
\begin{tabular}{c}-\end{tabular}&
\begin{tabular}{c}\textbf{Yes}\\(simulation)\end{tabular}\\
\hline
\begin{tabular}{c}\cite{brik2020federated}\end{tabular}&
\begin{tabular}{c}Single\\(UAVs)\end{tabular}&
\begin{tabular}{c}Single\\(wireless\\network\\data)\end{tabular}&
\begin{tabular}{c}Single\\(network\\management)\end{tabular}&
\begin{tabular}{c}Conventional\end{tabular}&
\begin{tabular}{c}No\end{tabular}&
\begin{tabular}{c}-\end{tabular}&
\begin{tabular}{c}No\end{tabular}\\
\hline
\begin{tabular}{c}\cite{li2019fc}\end{tabular}&
\begin{tabular}{c}Single\\(mobile\\robots)\end{tabular}&
\begin{tabular}{c}Single\\(visual)\end{tabular}&
\begin{tabular}{c}Single\\(SLAM)\end{tabular}&
\begin{tabular}{c}Conventional\end{tabular}&
\begin{tabular}{c}No\end{tabular}&
\begin{tabular}{c}-\end{tabular}&
\begin{tabular}{c}\textbf{Yes}\\(real-world,\\simulation)\end{tabular}\\
\hline
\begin{tabular}{c}\cite{bretan2019robot}\end{tabular}&
\begin{tabular}{c}Single\\(manipulation\\robots)\end{tabular}&
\begin{tabular}{c}Single\\(end-effector\\manipulation)\end{tabular}&
\begin{tabular}{c}Single\\(motion\\planning)\end{tabular}&
\begin{tabular}{c}Reinforcement\\learning\end{tabular}&
\begin{tabular}{c}No\end{tabular}&
\begin{tabular}{c}\textbf{Collaborative}\\\textbf{training}\end{tabular}&
\begin{tabular}{c}\textbf{Yes}\\(real-world,\\simulation)\end{tabular}\\
\hline
\begin{tabular}{c}\cite{zhou2018real}\end{tabular}&
\begin{tabular}{c}Single\\(perception\\robots)\end{tabular}&
\begin{tabular}{c}Single\\(visual)\end{tabular}&
\begin{tabular}{c}Single\\(visual\\perception)\end{tabular}&
\begin{tabular}{c}Conventional\end{tabular}&
\begin{tabular}{c}No\end{tabular}&
\begin{tabular}{c}-\end{tabular}&
\begin{tabular}{c}\textbf{Yes}\\(real-world,\\simulation)\end{tabular}\\
\hline
\begin{tabular}{c}\cite{majcherczyk2020flow}\end{tabular}&
\begin{tabular}{c}Single\\(mobile\\robots)\end{tabular}&
\begin{tabular}{c}Single\\(position)\end{tabular}&
\begin{tabular}{c}Single\\(trajectory\\forecasting)\end{tabular}&
\begin{tabular}{c}Conventional\\(also\\serverless\\variant)\end{tabular}&
\begin{tabular}{c}No\end{tabular}&
\begin{tabular}{c}-\end{tabular}&
\begin{tabular}{c}\textbf{Yes}\\(simulation)\end{tabular}\\
\hline
\begin{tabular}{c}\cite{imteaj2020fedar}\end{tabular}&
\begin{tabular}{c}Single\\(mobile\\robots)\end{tabular}&
\begin{tabular}{c}Single\\(visual)\end{tabular}&
\begin{tabular}{c}Single\\(visual\\perception)\end{tabular}&
\begin{tabular}{c}Conventional\\(resource-\\aware)\end{tabular}&
\begin{tabular}{c}No\end{tabular}&
\begin{tabular}{c}-\end{tabular}&
\begin{tabular}{c}\textbf{Yes}\\(simulation)\end{tabular}\\
\hline
\begin{tabular}{c}\cite{ng2020joint}\end{tabular}&
\begin{tabular}{c}Single\\(UAVs)\end{tabular}&
\begin{tabular}{c}Single\\(wireless\\network\\data)\end{tabular}&
\begin{tabular}{c}Single\\(IoV operation)\end{tabular}&
\begin{tabular}{c}Conventional\\(auction-\\coalition\\formation\\enhanced)\end{tabular}&
\begin{tabular}{c}No\end{tabular}&
\begin{tabular}{c}-\end{tabular}&
\begin{tabular}{c}\textbf{Yes}\\(real-world)\end{tabular}\\
\hline
\begin{tabular}{c}\cite{ferrer2018robochain}\end{tabular}&
\begin{tabular}{c}Single\\(humanoid\\robots)\end{tabular}&
\begin{tabular}{c}\textbf{Multiple}\\(audio,\\visual)\end{tabular}&
\begin{tabular}{c}Single\\(HRI)\end{tabular}&
\begin{tabular}{c}Conventional\\(blockchain\\enhanced)\end{tabular}&
\begin{tabular}{c}No\end{tabular}&
\begin{tabular}{c}-\end{tabular}&
\begin{tabular}{c}No\end{tabular}\\
\hline
\begin{tabular}{c}\cite{lim2020towards}\end{tabular}&
\begin{tabular}{c}Single\\(UAVs)\end{tabular}&
\begin{tabular}{c}Single\\(navigation\\sensing\\data)\end{tabular}&
\begin{tabular}{c}Single\\(IoV\\applications)\end{tabular}&
\begin{tabular}{c}Conventional\\(contract-\\matching\\enhanced)\end{tabular}&
\begin{tabular}{c}No\end{tabular}&
\begin{tabular}{c}-\end{tabular}&
\begin{tabular}{c}\textbf{Yes}\\(simulation)\end{tabular}\\
\hline\hline
\begin{tabular}{c}Proposed\end{tabular}&
\begin{tabular}{c}\textbf{Multiple}\\(arm, AGV,\\humanoid, UAV,\\vehicle, industrial,\\etc.)\end{tabular}&
\begin{tabular}{c}\textbf{Multiple}\\(vision, light,\\force, acoustic,\\motion, pressure,\\etc.)\end{tabular}&
\begin{tabular}{c}\textbf{Multiple}\\(sensing, navigation,\\manipulation, control,\\HRI, etc.)\end{tabular}&
\begin{tabular}{c}\textbf{Multiple}\\(reinforcement learning,\\apprenticeship learning,\\active learning,\\structured predictions,\\etc.)\end{tabular}&
\begin{tabular}{c}\textbf{Yes}\\(user weighting,\\parameter weighting,\\user clustering)\end{tabular}&
\begin{tabular}{c}\textbf{Transfer}\\\textbf{learning},\\\textbf{multi-task}\\\textbf{learning},\\\textbf{meta}\\\textbf{learning}\end{tabular}&
\begin{tabular}{c}No\end{tabular}\\
\hline
\end{tabular}
\end{center}
\label{t:cognitiveArchitectures}
\end{table*}

\begin{table*}[t]
\caption{Reported performance of multi-agent FL-based robotic systems of the literature}
\setlength\tabcolsep{1pt}
\begin{center}
\begin{tabular}{|c||c|c|c|}
\hline
&\multicolumn{3}{c|}{\begin{tabular}{c}Evaluation framework\end{tabular}}\\
\cline{2-4}
\centering\begin{tabular}{c}Method\end{tabular}&
\begin{tabular}{c}Robotic task\end{tabular}&
\begin{tabular}{c}Experimental setting\end{tabular}&
\begin{tabular}{c}Reported performance\end{tabular}\\
\hline
\hline
\begin{tabular}{c}\cite{Liu19lifelong}\end{tabular}&
\begin{tabular}{p{3cm}}Navigation in cloud robotic systems\end{tabular}&
\begin{tabular}{p{7cm}}Use of 4 virtual indoor environments (containing static obstacles, such as cardboard boxes, dustbins, cans, etc.) simulated in Gazebo, using Turtlebot3 as the test platform.\\Real-world experiment using Turtlebot3 platform in an indoor office environment.\end{tabular}&
\begin{tabular}{p{6cm}}The robot successfully navigates in an indoor office environment in an automatic way.\end{tabular}\\
\hline
\begin{tabular}{c}\cite{liang2019federated}\end{tabular}&
\begin{tabular}{p{3cm}}Autonomous car collision avoidance system \end{tabular}&
\begin{tabular}{p{7cm}}Use of 3 real-world JetsonTX2 remote autonomous vehicles, lidar sensor used to only collect distance data from the front view, use of Microsoft AirSim simulator, indoor environment with obstacle objects.
\end{tabular}&
\begin{tabular}{p{6cm}}The FL scheme leads to an increase of $27.2\%$ in the average distance from obstacles and a decrease of $42.5\%$ in collision counts, compared to single-car learning.\end{tabular}\\
\hline
\begin{tabular}{c}\cite{liu2020federated}\end{tabular}&
\begin{tabular}{p{3cm}}Self-driving cars in cloud robotic systems\end{tabular}&
\begin{tabular}{p{7cm}}Use of Microsoft AirSim and CARLA simulators, three different neighbourhood environments, three types of sensor data (RGB images, depth maps, semantic segmentation masks), six policy networks evaluated.\end{tabular}&
\begin{tabular}{p{6cm}}The FL scheme leads to performance improvement (e.g. avoiding collisions, making timely turns, etc.) of up to $17.4\%$ for different weather conditions (namely normal, rain, snow, fog and dust).\end{tabular}\\
\hline
\begin{tabular}{c}\cite{khamidehi2020federated}\end{tabular}&
\begin{tabular}{p{3cm}}UAVs' path planning, satisfying probabilistic connectivity constraint\end{tabular}&
\begin{tabular}{p{7cm}}Consideration of a 10x10km$^2$ area, 100m UAVs' operational altitude, a total of 10 collaborating UAVs.\end{tabular}&
\begin{tabular}{p{6cm}}Estimated UAV trajectories for three pairs of initial and final locations.\end{tabular}\\
\hline
\begin{tabular}{c}\cite{li2019fc}\end{tabular}&
\begin{tabular}{p{3cm}}SLAM in cloud robotic systems\end{tabular}&
\begin{tabular}{p{7cm}}Use of 2680 indoor scenes from the TUM dataset and 3 turtlebot2 platforms equipped with lidar and RGB camera in the Gazebo simulator.\end{tabular}&
\begin{tabular}{p{6cm}}The place mismatching rate (per image) exhibits values in the interval $[0.3,1.0]$, most of them in the range $[0.4,0.6]$.\end{tabular}\\
\hline
\begin{tabular}{c}\cite{bretan2019robot}\end{tabular}&
\begin{tabular}{p{3cm}}Motion planning in inverse kinematics, control and task planning\end{tabular}&
\begin{tabular}{p{7cm}}Forward kinematics: Prediction of 3D coordinates of a Rethink Robotics’ Sawyer Robot, 10,000 test samples.\\
Inverse kinematics: Use of a Sawyer Robot for generating joint angles that satisfy end effector coordinates, 100 test samples.\\
Nonlinear control: Use of the OpenAI gym pendulum-v0 environment with the task of keeping a friction-less pendulum standing up, 25 sequential actions considered.\\
Task planning: Generation of a sequence of drumstick trajectories, allowing it to strike the drum in a rhythmically controlled manner.\end{tabular}&
\begin{tabular}{p{6cm}}Forward kinematics: Mean Euclidean distance error (cm) in the range of $[3.57,3.69]$.\\
Inverse kinematics: Mean Euclidean distance (cm) in the range $[4.1,6.4]$. \\
Nonlinear control: Mean reward value in the range $[-141.23,-143.22]$.\\
Task planning: Cosine similarity equal to $1$ in simulation and $0.86$ in real-world settings.\end{tabular}\\
\hline
\begin{tabular}{c}\cite{zhou2018real}\end{tabular}&
\begin{tabular}{p{3cm}}Visual perception for handwritten digit recognition and object detection\end{tabular}&
\begin{tabular}{p{7cm}}Use of 100 simulation robots and two public image datasets (MNIST, CIFAR-10).\end{tabular}&
\begin{tabular}{p{6cm}}In CIFAR, accuracy up to $80\%$ achieved. In MNIST, accuracy up to more than $90\%$ accomplished.\end{tabular}\\
\hline
\begin{tabular}{c}\cite{majcherczyk2020flow}\end{tabular}&
\begin{tabular}{p{3cm}}Trajectory forecasting in a multi-agent setting\end{tabular}&
\begin{tabular}{p{7cm}}Generation of artificial navigation data in four multi-robot settings, use of 15 and 60 robot swarms, 10s trajectory length.\end{tabular}&
\begin{tabular}{p{6cm}}Average displacement error (cm) in the area of $0.21$ when 60 robots involved, respective final displacement error value equal to $0.11$.\end{tabular}\\
\hline
\begin{tabular}{c}\cite{imteaj2020fedar}\end{tabular}&
\begin{tabular}{p{3cm}}Visual perception for digit images recognition\end{tabular}&
\begin{tabular}{p{7cm}}Consideration of 12 mobile robots and a combination of MNIST digit classification dataset with an own captured one.\end{tabular}&
\begin{tabular}{p{6cm}}Visual recognition accuracy up to more than $90\%$.\end{tabular}\\
\hline
\begin{tabular}{c}\cite{ng2020joint}\end{tabular}&
\begin{tabular}{p{3cm}}IoV operation, e.g. for traffic prediction\end{tabular}&
\begin{tabular}{p{7cm}}Use of a Cartesian grid of size 1000×1000 and vehicles (as the IoV components) that communicate with UAVs in simulation settings.\end{tabular}&
\begin{tabular}{p{6cm}}Over $10\%$ relative improvement in profit-maximizing behavior for reinforcing the IoV operations.\end{tabular}\\
\hline
\begin{tabular}{c}\cite{lim2020towards}\end{tabular}&
\begin{tabular}{p{3cm}}IoV applications, e.g. traffic prediction and car park occupancy management
\end{tabular}&
\begin{tabular}{p{7cm}}Different combinations of up to 7 UAVs and 6 subregions.
\end{tabular}&
\begin{tabular}{p{6cm}}Simulation results for the defined sensing case studies.
\end{tabular}\\
\hline
\end{tabular}
\end{center}
\label{t:FLPerformance}
\end{table*}

The staggering technological capabilities and tremendous innovation potentials that have recently been introduced by FL techniques in multiple AI and data science application fields inevitably leads to transformative developments in the robotics area as well. Indeed, very lately an initial body of works investigating relatively straight-forward implementations of FL schemes in specific robotic tasks has been presented and focus on the following main aspects:
\begin{itemize}
\item Autonomous navigation: Liu et al. \cite{Liu19lifelong} introduce a reinforcement learning approach, coupled with suitable knowledge fusion and transfer learning algorithms, for autonomous navigation of mobile robots using a cloud environment. Additionally, a reinforcement learning-based real-life collision avoidance system for indoor settings with obstacle objects is presented in \cite{liang2019federated}. Majcherczyk et al. \cite{majcherczyk2020flow} investigate a multi-agent trajectory forecasting problem, following both a conventional and a serverless FL aggregation mechanism. Moreover, an imitation learning framework is proposed for generating guidance models for robots in a self-driving task application in \cite{liu2020federated}. Furthermore, different approaches have also been investigated for the case of Unmanned Aerial Vehicles (UAVs) path planning \cite{khamidehi2020federated}\cite{brik2020federated}.
\item Simultaneous Localization And Mapping (SLAM): Li et al. \cite{li2019fc} present a visual-lidar SLAM approach, which supports feature extraction and dynamic vocabulary designation in real-time, while operating on a cloud workstation.
\item Motion planning: Bretan et al. \cite{bretan2019robot} incorporate principles of ensemble and reinforcement learning, as well as gradient free optimization, for various robotic tasks, including inverse kinematics, controls and planning.
\item Visual perception: Zhou et al. \cite{zhou2018real} examine a differential privacy protection approach to multiple robotic recognition tasks, while balancing the trade-off between performance and privacy. Additionally, a resource-aware learning approach for distributed mobile robots, taking into account limitations either in memory, bandwidth, processor or battery life, is presented in \cite{imteaj2020fedar}.
\item Vehicle management: Ng et al. \cite{ng2020joint} propose the use of UAVs as wireless relays to facilitate the communication between Internet of Vehicles (IoV) components and the FL server, targeting to improve the accuracy of the FL framework. Additionally, a variant of the previous work focuses on the deployment of a privacy-preserving federation of Drones-as-a-Service (DaaS) providers for the development of IoV applications, e.g. traffic prediction and car park occupancy management \cite{lim2020towards}.
\item Human-robot interaction: Ferrer et al. \cite{ferrer2018robochain} introduce a multi-robot framework in the domain of mobile health, targeting to facilitate clinical interventions by improving the robots' interaction capabilities.
\end{itemize}

Taking into account the above analysis, it can be seen that the relevant robotics literature has so far focused on relatively straight-forward implementations of the FL paradigm. In particular, Table \ref{t:cognitiveArchitectures} illustrates the main architectural characteristics of current multi-agent FL-based robotic systems, emphasizing on the following crucial aspects: a) the types of robots involved, b) the types of sensors used, c) the types of tasks modeled, d) the actual learning methodology implemented (under the decentralized FL setting), e) if the human factor is integrated in the learning loop, f) whether any additional advanced learning methodologies are incorporated and g) if the cognitive architecture has been implemented. The proposed multi-agent LfD cognitive AI architecture (to be detailed in Section \ref{sec:CognitiveArchitecture}) is also included in Table \ref{t:cognitiveArchitectures}, where it can be seen that it achieves to introduce the theoretical foundations along with specific implementation$/$algorithmic guidelines for widening the capabilities of current systems in all examined aspects. More specifically, the introduced cognitive AI architecture is particularly beneficial for addressing the following key limitations in the literature of critical importance: 
\begin{itemize}
\item The available methods examine only individual robotic tasks in an isolated way, while in typical real-world applications multiple tasks, as well as their cross-correlations, should be simultaneously examined; on the contrary, the proposed architecture aims at jointly learning multiple and significantly diverse robotic tasks (e.g. sensing, navigation, manipulation, control, HRI, etc.).
\item Current approaches are only constrained in self-learning scenarios; however, the introduced architecture investigates sophisticated AI-empowered HRI schemes and, hence, involves the human in the learning loop for exhibiting numerous advantageous characteristics, especially for demanding and fine-grained robotic tasks (e.g. more precise guidance, more time efficient learning, inspection of learning procedure, etc.). Towards this direction, different techniques for incorporating multi-user feedback information are provided (namely user weighting, parameter weighting and user clustering).
\item Literature works usually do not take advantage of advanced AI-enabled learning methodologies for strengthening their learning capabilities; on the other hand, the proposed architecture incorporates different such methodologies (namely transfer, multi-task and meta learning), particularly focusing on exploiting cross-task correlation information.
\end{itemize}

\begin{figure*}[th]
\centering
\includegraphics[width=0.85\textwidth]{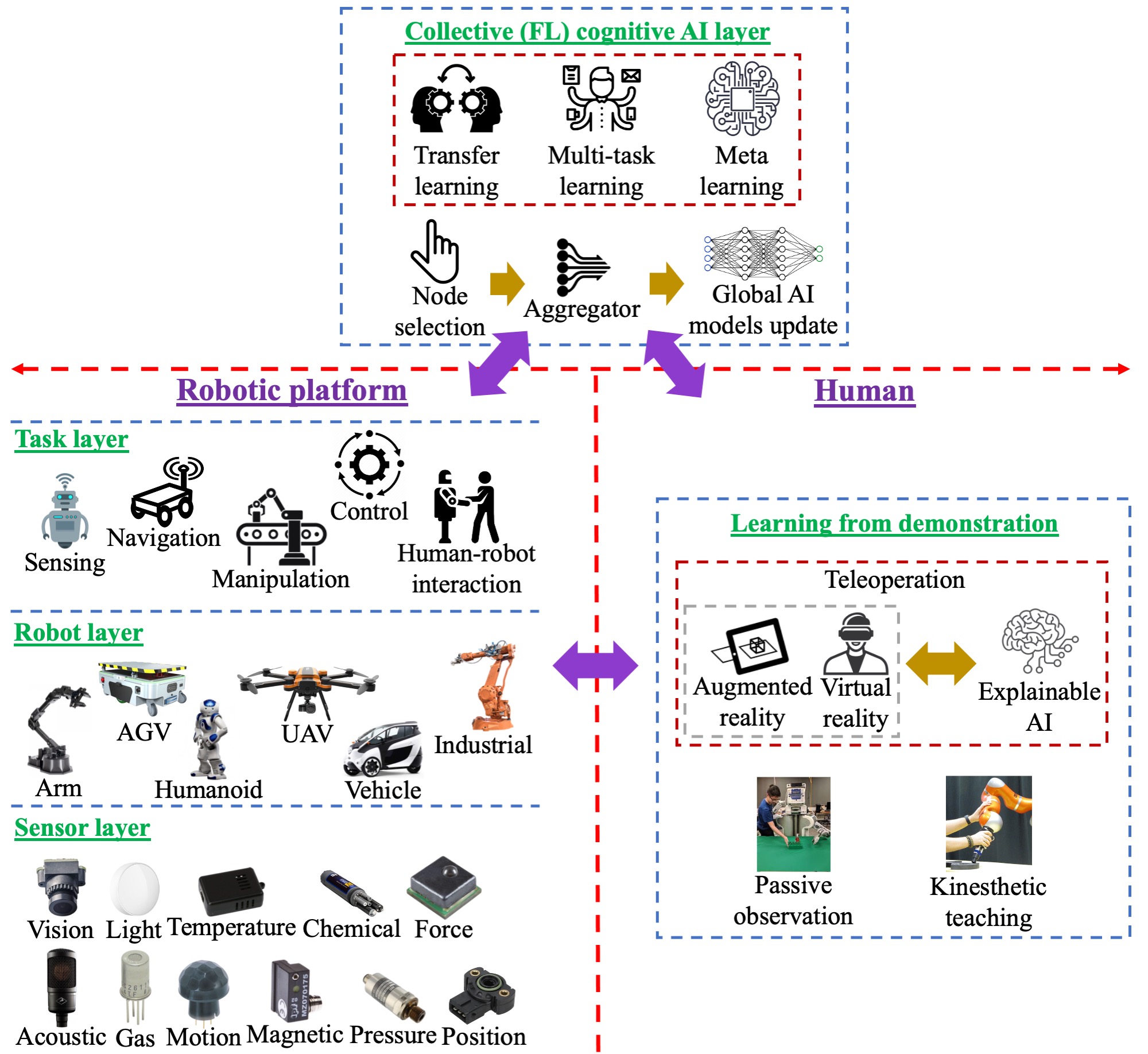}
\caption{High-level representation of the introduced multi-agent cognitive AI collaborative learning architecture.}
\label{f:FLenvironment}
\end{figure*}

In order to provide insights also with respect to performance aspects of current multi-agent FL-based robotic systems, Table \ref{t:FLPerformance} outlines evaluation-related details of the architectures described in Table \ref{t:cognitiveArchitectures} that have been implemented. In particular, the key points of the evaluation framework developed in each study are provided, focusing on presenting: a) the robotic tasks to be reinforced using the FL paradigm, b) the experimental setting used during the evaluation process and c) the reported performance measurements (in simulation and$/$or real-world environments). From the provided results, it can be observed that FL technologies lead to promising outcomes and significant potentials in various application cases. However, more comprehensive, detailed and thorough evaluation (also in additional robotic tasks and well-defined challenging open benchmarks) is required in order to demonstrate in a robust way the benefits of (multiple and diverse aspects of) FL in practical applications.

\section{Multi-agent LfD cognitive AI architecture}
\label{sec:CognitiveArchitecture}

In this section, a novel open expandable cognitive AI-empowered architecture for multi-agent human-robot collaborative learning is introduced. The ultimate goal is to provide reliable solutions to current critical challenges faced by robotic systems (and especially within the particular LfD field), namely those of multi-agent learning \cite{Hussein17} and long-term autonomy \cite{Kunze18}; achieving the latter will in turn facilitate the deployment and wide-spread use of robots in open and complex environments. The fundamental conceptualization behind the designed architecture is to leverage the recent technological advancements in the field of AI (focusing mainly on the use of the FL paradigm and closely related technologies) and to transfer them to the robotics application area for extending the capabilities of the current techniques. The designed cognitive AI architecture, whose high-level representation is illustrated in Fig. \ref{f:FLenvironment}, incarnates a multi-human multi-robot collaborative learning environment that is composed of the following main reference entities: a) the robotic platform, b) the human, and c) the collective (FL) cognitive AI layer. Detailed analysis of the formalisms and roles of the aforementioned entities are provided in the remaining of the section, while a summary of the main mathematical symbols used is given in Table \ref{t:symbols}. 

\subsection{Robotic platform}
\label{ssec:RoboticPlatform}

Under the proposed conceptualization, each robotic platform consists of the following main layers: a) the sensor, b) the robot, and c) the task one. In particular, the sensor layer includes the set of sensing devices that allow the robotic platform to perceive and to collect critical information about the surrounding environment. The set of potentially supported types of sensors, which can be significantly broad and also depends on the particular application case, is defined as follows:
\begin{eqnarray}\nonumber
S=\{s_i | i\in[1,I], i\in\mathbb{N}, I\in\mathbb{N}   \}\\\nonumber
=\{Vision, Light, Temperature, Chemical, Force,\\\nonumber
Acoustic, Gas, Motion, Magnetic, Pressure,\\
Position, ...   \}
\end{eqnarray} 
Regarding the robot layer, this refers to the actual mechatronic equipment to be deployed. Depending on the specific operational scenario, multiple types of robots can be used, supporting different requirements for mobility, positioning, manipulation, communication, size, payload, etc. The set of available types of robots is denoted:
\begin{eqnarray}\nonumber
R=\{r_j | j\in[1,J], j\in\mathbb{N}, J\in\mathbb{N}   \}\\\nonumber
=\{Arm, AGV, Humanoid, UAV,\\
Vehicle, Industrial, ...   \}
\end{eqnarray} 
With respect to the task layer, this concerns the types of policies and activities that the robotic platform will be required to implement. These may cover a large set of possible perception, cognition, motor and interaction functionalities, which will inevitably need to be appropriately adapted, apart from the specific application requirements at hand, also to the particularities of the employed robot type $r_j$. The set of potential types of robotic tasks is defined as follows:
\begin{eqnarray}\nonumber
T=\{t_k | k\in[1,K], k\in\mathbb{N}, K\in\mathbb{N}   \}\\\nonumber
=\{Sensing, Navigation, Manipulation,\\
Control, Human-robot~interaction, ...   \}
\end{eqnarray} 
Taking into account the above-mentioned formalisms, a robotic platform $P_l$ can be fully specified as follows:
\begin{eqnarray}\label{eq:roboticPlatform}
P_l=\{ S_l, R_l, T_l | S_l \subseteq S, R_l \subseteq R, T_l \subseteq T \}, l\in[1,L],
\end{eqnarray} 
where $L$ denotes the total number of robotic platforms present in the examined cognitive environment.

\subsection{Human}
\label{ssec:Human}

\begin{table}[t]
\caption{List of main mathematical symbols}
\begin{center}
\begin{tabular}{|c|l|}
\hline
\begin{tabular}{l}Symbol\end{tabular}&\begin{tabular}{l}Description\end{tabular}\\
\hline\hline
\begin{tabular}{l}$S=\{s_i | i\in[1,I]\}$\end{tabular}&\begin{tabular}{l}Set of sensor types\end{tabular}\\\hline
\begin{tabular}{l}$R=\{r_j | j\in[1,J]\}$\end{tabular}&\begin{tabular}{l}Set of robot types\end{tabular}\\\hline
\begin{tabular}{l}$T=\{t_k | k\in[1,K]\}$\end{tabular}&\begin{tabular}{l}Set of robotic task types\end{tabular}\\\hline
\begin{tabular}{l}$P_l,~l\in[1,L]$\end{tabular}&\begin{tabular}{l}Robotic platform\\(also FL network node)\end{tabular}\\\hline
\begin{tabular}{l}$H=\{h_m | m\in[1,M]\}$\end{tabular}&\begin{tabular}{l}Set of human operators\end{tabular}\\\hline
\begin{tabular}{l}$\textbf{W}_\beta,~\beta\in[1,B]$\end{tabular}&\begin{tabular}{l}Global AI model\end{tabular}\\\hline
\begin{tabular}{l}$\textbf{W}_\beta^{l,m}$\end{tabular}&\begin{tabular}{l}Local version of AI model $\textbf{W}_\beta$ for\\human $h_m$ at node $P_l$\end{tabular}\\\hline
\begin{tabular}{l}$\mathcal{L}_\beta$\end{tabular}&\begin{tabular}{l}Loss function defined for $\textbf{W}_\beta$\end{tabular}\\\hline
\begin{tabular}{l}$\delta^l$\end{tabular}&\begin{tabular}{l}Local dataset stored at node $P_l$\end{tabular}\\\hline
\begin{tabular}{l}$\textbf{Q}^{l,m}$\end{tabular}&\begin{tabular}{l}Local profile of human user $h_m$\\at node $P_l$\end{tabular}\\\hline
\begin{tabular}{l}$\textbf{Q}^{g,m}$\end{tabular}&\begin{tabular}{l}Global profile of human user $h_m$\end{tabular}\\\hline
\begin{tabular}{l}$\mathcal{L}_g$\end{tabular}&\begin{tabular}{l}Global loss function,\\considering all $\textbf{W}_\beta$\end{tabular}\\\hline
\end{tabular}
\end{center}
\label{t:symbols}
\end{table}

Within the designed multi-agent collaborative learning environment, the human factor constitutes a fundamental building block for simultaneously a) transferring fine-grained and sophisticated skills to the robot and b) supervising/inspecting the learned robot behaviors (i.e. in principle guiding the overall robotic learning process). Although all types of LfD methodologies (namely kinesthetic teaching, teleoperation and passive observation, as detailed in Section \ref{sec:Introduction}) are supported by the introduced cognitive AI architecture, the adoption of a teleoperation-based approach is considered to exhibit significant advantageous characteristics. In particular, an intuitive and sophisticated teleoperation scheme is foreseen that is based on the combined used of Augmented Reality (AR) visualization mechanisms and eXplainable AI (XAI) technologies.

Regarding AR techniques, they have so far been shown to be beneficial for enabling the human operators to be simultaneously aware of the actual processes that take place in the physical environment (e.g. a factory) and to be continuously updated with valuable information related to the underlying automatic control procedures. In the current cognitive architecture, AR tools are employed in order to allow the human user to perform a physical inspection of the robot exhibited behaviors (and hence to identify malfunctions, hazardous situations, deviations from desired policies, etc.), while at the same time being constantly provided with key detailed insights about the AI processes being applied (e.g. the AI modules being used, their estimated outputs, how specific decisions are reached, etc.). To this end, this AR-grounded setting constitutes an efficient and user-friendly way to examine the convergence of the physical (robot) and AI (software) worlds at the same time. Additionally, while also of paramount importance, the designed AR setting makes two key functionalities available to the human user: a) to provide feedback regarding the possible corrections in the exhibited robot behaviors (when a deviation from the desired targets is observed) and b) to receive full control of the robot performed actions, through a teleoperation-based robot policy definition scheme. Under certain circumstances (e.g. involvement of very large-scale application settings, like industrial operating plants), Virtual Reality (VR) technologies can also be used in conjunction (e.g. virtual factory) for overall process monitoring. It needs to be mentioned that the set of human operators involved in the designed cognitive architecture is denoted $H=\{h_m | m\in[1,M], m\in\mathbb{N}, M\in\mathbb{N}\}$.

Concerning XAI methods, their fundamental aim is to improve trust and transparency of AI-based systems, by attempting to provide valuable insights or to directly explain the decision making process of AI procedures \cite{adadi2018peeking}. Under the current conceptualization, XAI techniques are adopted in order to provide precise explanations/insights to the human operator regarding the deviation of the robot behavior from the desired one, i.e. enabling an in depth inspection of the robot behavior. The XAI-based generated explanations are provided to the human user through the aforementioned AR/VR visualization interfaces. Consequently, the human user is capable of providing more accurate guidance to the robot (through means of provision of feedback or teleoperation, as discussed above) and, hence, to supervise the overall robotic learning process more closely. Depending on the particular type of sensor $s_i$ and task $t_k$, different model-specific or model-agnostic XAI methods can be employed \cite{arrieta2020explainable}.

\subsection{Collective (FL) cognitive AI layer}
\label{ssec:FLLayer}

The introduced collective (FL) cognitive AI layer constitutes the core entity in the designed architecture that is responsible for creating, updating and distributing multiple AI modules, which underpin the various robotic processes implemented in the examined application setting. These AI modules are collectively created and maintained by simultaneously aggregating knowledge/feedback from a large-scale human-robot interaction set (adopting the fundamental mechanisms of the LfD methodology), while aiming to address the current challenges of multi-agent learning and long-term autonomy. For achieving the latter goals, the designed open and expandable cognitive AI architecture is grounded on principles of the FL approach, whose fundamental mechanism is explained in Section \ref{sec:FL}. The main building blocks of the introduced cognitive AI layer, their formalism and detailed explanation of their functionalities, is provided in the followings:\\
\underline{{\it{Network nodes}}}: Every robotic platform $P_l$ corresponds to a network node of the defined architecture, in accordance to the fundamental FL mechanism illustrated in Fig. \ref{f:FL}. Each $P_l$ stores locally the generated data, which in principle contain information collected from the set of sensors $S_l$ incorporated by $P_l$. It needs to be highlighted that feedback information can be obtained by any $P_l$ from the interaction with any human subject $h_m$ present in the application environment, i.e. it is considered that any teacher $h_m$ can inspect and provide guidance to any performing robotic platform $P_l$.\\
\underline{\it{AI models}}: The designed cognitive architecture aims to address the needs related to the deployment of large-scale AI-driven human-robot environments, simultaneously supporting multiple combinations of sensor $s_i$, robot $r_j$ and task $t_k$ types. For achieving that, a broad set of individual AI models denoted $\textbf{W}_\beta \leftarrow f_1(S_\beta, R_\beta, T_\beta)$, underpinning the various implemented cognitive processes present in the examined environment, is considered, where $S_\beta \subseteq S$, $R_\beta \subseteq R$, $T_\beta \subseteq T$, $\beta\in[1,B]$, $\beta\in\mathbb{N}$, $B\in\mathbb{N}$ and $f_1(.)$ implies a generalized function or process that defines the exact NN-based materialization of model $\textbf{W}_\beta$ while considering $S_\beta$, $R_\beta$ and $T_\beta$ as input parameters. The overall goal of the introduced FL-grounded environment is to construct global $\textbf{W}_\beta$ models (Section \ref{sec:FL}), by exploiting data from a large-scale human-robot interaction set of distributed sources, while each network node can maintain a local/customized version of $\textbf{W}_\beta$.\\
\underline{\it{Learning methodology}}: Regarding the specific methodology to be followed for refining the robot learned policies (i.e. for updating the AI models $\textbf{W}_\beta$), different options can be investigated (e.g. reinforcement learning, transfer learning, active learning, etc.), as discussed in Section \ref{sec:Introduction}. The most suitable selection depends on the particularities of the application domain and each individually examined $\textbf{W}_\beta$.\\
\underline{\it{Local parameter updates}}: Regardless of the particular learning methodology selected, each robotic platform $P_l$ can estimate updates for any AI model $\textbf{W}_\beta$ that is associated with, using its locally stored data. More specifically, the following local parameter update mechanism is applied:
\begin{eqnarray}\label{eq:localModels}
\textbf{W}_\beta^{l,m} \leftarrow \textbf{W}_\beta^{l,m} - lr^l  \cdot \nabla \mathcal{L}_\beta (\textbf{W}_\beta^{l,m}, \delta^l),
\end{eqnarray} 
where $\textbf{W}_\beta^{l,m}$ is the local/customized version of the global model $\textbf{W}_\beta$ with respect to human teacher $h_m$, $lr^l$ is the local learning rate, $\nabla$ denotes the gradient of a function, $\mathcal{L}_\beta(.)$ represents the loss function defined for $\textbf{W}_\beta$ and $\delta^l$ is the locally stored dataset. 
Consequently, the local parameter updates, which are iteratively estimated, to be sent to the central node (aggregator) are computed as follows:
\begin{eqnarray}\nonumber
\Delta \textbf{W}_\beta^{l,m} = \textbf{W}_\beta^{l,m} - \textbf{W}_\beta
\end{eqnarray} 
From the above formalization, it can be seen that the designed architecture allows local parameter updates to be estimated separately for each human teacher $h_m$, i.e. also leading to `personalized' versions of each model $\textbf{W}_\beta$.\\
{\underline{{\it{User profiling}}}: Incorporating human feedback constitutes a fundamental part of the LfD approach and, consequently, of the current cognitive architecture. However, the latter poses additional challenges to the problem formulation that need to be efficiently addressed (e.g. variance in human performance, variability in knowledge across human subjects, learning from noisy/imprecise human input, learning from very large or very sparse datasets, incremental learning, etc.). Towards this direction and in parallel with the $\textbf{W}_\beta^{l,m}$ estimation process, an individual user profile $\textbf{Q}^{l,m} \leftarrow f_2(\delta^l, S_m)$ is constructed for every human subject $h_m$ at every node $P_l$, given the appropriate sensorial data $S_m$ to model the observed human behavior and a generalized function $f_2(.)$ that defines the exact (NN-based) implementation of the user profile while considering $S_m$ as input parameters. The aim of $\textbf{Q}^{l,m}$ is to cover physical (e.g. human actions, physical capabilities, etc.), cognitive (e.g. intention, personality, etc.) and social (e.g. non-verbal cues, emotions, etc.) aspects, in order to model and efficiently interpret the exhibited human activity \cite{rossi2017user}. Under the current conceptualization, DL-based approaches are considered for generating $\textbf{Q}^{l,m}$, as in \cite{farnadi2018user} and \cite{liang2020drprofiling}, aiming at combining increased modelling capabilities and easier integration to the designed FL-based framework.\\
\underline{\it{Global model update}}: Having computed the local parameter updates $\Delta \textbf{W}_\beta^{l,m}$ and the corresponding user profiles $\textbf{Q}^{l,m}$ (using locally generated and processed data $\delta^l$), these are sent to the central node (aggregator) so as to periodically produce updated versions of the global $\textbf{W}_\beta$ models. According to the conventional FL mechanism, an updated version $\textbf{W}'_\beta$ of each individual $\textbf{W}_\beta$ is generated on a regular basis, by applying a simple average operator, as follows:
\begin{eqnarray}\label{eq:FLGoblaModels}
\textbf{W}'_\beta = \textbf{W}_\beta + lr^g  \cdot \boldsymbol{\Gamma}_\beta\\\label{eq:FLGoblaUpdate}
\boldsymbol{\Gamma}_\beta = \frac{1}{\Lambda_\beta} \sum_{(l,m)} \Delta \textbf{W}_\beta^{l,m},
\end{eqnarray} 
where $lr^g$ denotes the global learning rate and $\Lambda_\beta$ the total number of received $\Delta \textbf{W}_\beta^{l,m}$ updates. The strength of the above-mentioned mechanism lies on incorporating a very large number $\Lambda_\beta$ of samples and multiple iterative updates of $\textbf{W}'_\beta$ that will likely lead to the convergence to well-performing and robust $\textbf{W}_\beta$ models, while the network nodes $P_l$ contributing in (\ref{eq:FLGoblaUpdate}) may be sampled out of the available ones (usually in a random way). However, combining information ($\Delta \textbf{W}_\beta^{l,m}$) related to different human subjects $h_m$ (that presumably exhibit highly diverse and varying behavior) is in turn very likely to lead the FL process to be confined to a local maximum or even to a non-convergence of the FL procedure; hence, jeopardising the overall learning process.

\subsubsection{Incorporation of multi-user feedback information}
\label{sssec:UserIncorporation}

For robustly confronting the observed variance in the behavior of the large number of involved human teachers $h_m$, the designed cognitive architecture (apart from possible sensorial data $s_i$ pre-processing for invariance incorporation) encompasses the estimated user profiles $\textbf{Q}^{l,m}$ in the global model update process, modifying (\ref{eq:FLGoblaUpdate}) to the following general formalism:
\begin{eqnarray}\label{eq:FLGlobalUpdateHuman}
\boldsymbol{\Gamma}_\beta = \Phi (\{\textbf{Q}^{l,m}\},\{ \Delta \textbf{W}_\beta^{l,m}\}), 
\end{eqnarray} 
where $\Phi(.)$ denotes a generalized function that combines the available $\textbf{Q}^{l,m}$ and $\Delta \textbf{W}_\beta^{l,m}$, estimated at every network node $P_l$. Depending on the particularities of the selected application domain (e.g. supported $\textbf{W}_\beta$, $S$, $R$, $T$, etc.), the following main materializations of $\Phi(.)$, while also being possible to be combined, are considered:
\begin{itemize}

\item {\it{User weighting}}: Under this conceptualization, the contribution of each human teacher $h_m$ is modulated by a different weight factor based on his/her exhibited behavior, as follows:
\begin{eqnarray}\nonumber
\boldsymbol{\Gamma}_\beta = \frac{1}{E_\beta} \sum_{(l,m)} \varepsilon (\textbf{Q}^{l,m}) \cdot \Delta \textbf{W}_\beta^{l,m}\\\nonumber
\varepsilon (\textbf{Q}^{l,m})=  1 / \lVert \textbf{Q}^{g} - \textbf{Q}^{l,m} \rVert\\
E_\beta = \sum_{(l,m)} \varepsilon (\textbf{Q}^{l,m}),\label{eq:UserWeight}
\end{eqnarray} 
where $\varepsilon (\textbf{Q}^{l,m})$ denotes the weight factor for each $h_m$ at every $P_l$, $\textbf{Q}^{g}$ the corresponding global user model (e.g. estimated through the same FL-based mechanism used for constructing $\textbf{W}_\beta$) and $\lVert . \rVert$ a similarity score metric (e.g. Euclidean distance between the parameters of the involved user models).

\item {\it{Parameter weighting}}: The fundamental consideration lies on performing sensitivity analysis \cite{zhang2015sensitivity}\cite{shu2019sensitivity} for estimating the degree of correlation among the parameters of $\textbf{W}_\beta^{l,m}$ and $\textbf{Q}^{l,m}$, i.e. emphasizing on how the exhibited behavior of each user $h_m$ affects individual aspects/parameters of $\textbf{W}_\beta^{l,m}$, according to the following formalism:
\begin{eqnarray}\nonumber
\boldsymbol{\Gamma}_\beta = \frac{1}{\textbf{R}_\beta} \sum_{(l,m)} \textbf{r}^{l,m}_\beta \cdot \Delta \textbf{W}_\beta^{l,m}\\\nonumber
\textbf{r}^{l,m}_\beta = U (\textbf{Q}^{l,m}, \textbf{W}_\beta^{l,m})\\\label{eq:FLGlobalUpdateParametersWeighting}
\textbf{R}_\beta = \sum_{(l,m)} \textbf{r}^{l,m}_\beta,
\end{eqnarray} 
where function $U(.)$ realizes sensitivity analysis for estimating the impact of parameters $\textbf{Q}^{l,m}$ on the respective ones of $\textbf{W}_\beta^{l,m}$, while matrix $\textbf{r}^{l,m}_\beta$ summarizes the estimated correlations.

\item {\it{User clustering}}: The main principle behind this approach, the so called `Multi-Center Federated Learning' \cite{xie2020multi}, considers the generation of multiple instances $\textbf{W}_{\beta,\eta}$ of each global model $\textbf{W}_\beta$, in order to better capture the heterogeneity of data distributions across different users. In particular, each human teacher $h_m$ at each local network node $P_l$ is associated with a single $\textbf{W}_{\beta,\eta}$; the latter is iteratively updated, by elaborating (\ref{eq:FLGlobalUpdateHuman}), as follows:
\begin{eqnarray}\label{eq:FLGlobalUpdateUserClustering}
\boldsymbol{\Gamma}_\beta = \frac{\sum_{(l,m)} \theta^{l,m}_{\beta,\eta} \Delta \textbf{W}_\beta^{l,m}}{\sum_{(l,m)} \theta^{l,m}_{\beta,\eta}},
\end{eqnarray} 
where $\theta^{l,m}_{\beta,\eta}=1$, if teacher $h_m$ at node $P_l$ is associated with $\textbf{W}_{\beta,\eta}$, and $\theta^{l,m}_{\beta,\eta}=0$, otherwise.
\end{itemize}
It needs to be highlighted that all above-mentioned variants of function $\Phi(.)$ in (\ref{eq:UserWeight})-(\ref{eq:FLGlobalUpdateUserClustering}) are considered to be implemented in a neural-network form; hence, rendering the overall approach end-to-end learnable within the same integrated FL scheme.

\subsubsection{Incorporation of cross-task correlation information}
\label{sssec:TaskIncorporation}

Complementary to the integration of information from multiple human teachers $h_m$ (as detailed in Section \ref{sssec:UserIncorporation}), the designed cognitive layer puts also particular emphasis on analysing, modelling and exploiting the correlations among the multiple (and often co-occurring) robotic tasks, which are controlled by the AI models $\textbf{W}_\beta$. Towards this direction, the aforementioned FL-based mechanisms for incorporating multi-human feedback information is further elaborated and enhanced, by integrating the following sophisticated AI-empowered learning capabilities: 

\begin{itemize}

\item {\it{Transfer learning}}: Federated transfer learning constitutes a suitable methodology for addressing cases where different AI-driven processes share an overlap in the respective feature space, aiming at exploiting the underlying data correlations and building models collaboratively \cite{liu2018secure}\cite{chen2020fedhealth}. In particular, the conventional FL mechanism (described in Section \ref{sssec:UserIncorporation}) considers the separate construction of each global $\textbf{W}_\beta$ model, using the locally generated data $\delta^l$, by adopting the following general type of loss function during the training step:
\begin{eqnarray}\label{eq:FLGlobalLossFL}
\mathcal{L}_g =\sum_{\beta, (l,m)} \mathcal{L}_{\beta} ( \hat{Y} (\textbf{W}_\beta), Y (\textbf{W}_\beta))
\end{eqnarray} 
where $\mathcal{L}_g$ denotes the employed global loss function, $\mathcal{L}_{\beta}$ corresponds to the loss term with respect to each individual $\textbf{W}_\beta$ and $\hat{Y}(.)$, $Y(.)$ is the estimated, targeted (ground truth) output of $\textbf{W}_\beta$, respectively. In order to achieve feature transfer learning, the following alignment loss factor is added to $\mathcal{L}_g$ in (\ref{eq:FLGlobalLossFL}):
\begin{eqnarray}\label{eq:FLGlobalLossFLTransfer}
\mathcal{L}_2 =\sum_{(\beta 1, \beta 2), (l,m)} V ( \textbf{W}_{\beta 1}, \textbf{W}_{\beta 2})
\end{eqnarray} 
where $V(.)$ denotes an alignment/similarity measure between $\textbf{W}_{\beta 1}$ and $\textbf{W}_{\beta 2}$ (e.g. Euclidean distance $\lVert . \rVert$ between the model parameters, as in (\ref{eq:UserWeight})). It needs to be highlighted that the above mentioned transfer learning mechanism is applicable for AI models that exhibit similar patterns and correlations in the underlying data space, which inevitably implies overlaps and similarities between the respective $S_{\beta 1}$ and $S_{\beta 2}$ sets. Moreover, the principal goal of this scheme lies on transferring knowledge gained during the training process of model $\textbf{W}_{\beta 1}$ (where sufficiently large training datasets are available) to facilitate the creation of a functionally similar/related model $\textbf{W}_{\beta 2}$ (where availability of adequate amounts of training data is not present).

\item {\it{Multi-task learning}}: According to the formalisms proposed in the literature so far for federated Multi-Task Learning (MTL), the fundamental aim is posed as simultaneously constructing separate, but related, AI models at each network node \cite{smith2017federated}\cite{sattler2020clustered}. The latter requires, among others, the formulation of a so called precision matrix that encodes the inter-relations among the models, which can be either explicitly defined a priori or learned directly from the data. Under the current conceptualization, the problem of federated multi-task learning is re-formulated so as to allow the simultaneous learning of multiple global models $\textbf{W}_\beta$ that correspond to related (and often co-occurring) robotic tasks. The respective loss function to be used during training has the following general form:
\begin{eqnarray}\nonumber
\mathcal{L}_g = \sum_{\beta, (l,m)} \mathcal{L}_{\beta} ( \hat{Y} (\textbf{W}_\beta), Y (\textbf{W}_\beta))+ \mathcal{X} (\mathbf{A}, \mathbf{\Omega})\\
\mathbf{A}=[\textbf{W}_1~\textbf{W}_2~ ...~ \textbf{W}_B],\label{eq:MultiTaskLearningLoss}
\end{eqnarray} 
where $\mathbf{A}$ is a weight matrix produced by the concatenation of the individual $\textbf{W}_\beta$ model parameters and $\mathbf{\Omega} \in\mathbb{R}^{B\times B}$ is the so called precision matrix. Function $\mathcal{X} (.)$ summarizes the defined assumptions of the federated multi-task learning problem, where a bi-convex formulation is often selected \cite{caldas2018federated}, as follows:
\begin{eqnarray}\nonumber
\mathcal{X} (\mathbf{A}, \mathbf{\Omega})=\frac{\lambda}{2}tr(\mathbf{A}\mathbf{\Omega}\mathbf{A}^T)\\
\mathbf{\Omega}^{-1}\geq0,~tr(\mathbf{\Omega}^{-1})=1,\label{eq:MultiTaskLearning}
\end{eqnarray} 
where $\lambda$ is a constant and $tr(.)$ denotes the trace of a matrix. The ultimate goal of MTL is to jointly construct multiple robust $\textbf{W}_\beta$ models (by taking advantage of the cross-correlations among the respective robotic tasks), rather than only transplanting collected knowledge between different models (as the transfer learning scheme in principle does). Moreover, the multi-task learning formalism outlined in (\ref{eq:MultiTaskLearningLoss}) and (\ref{eq:MultiTaskLearning}) corresponds to the most commonly met type of MTL, which often assumes that the involved $\textbf{W}_\beta$ models operate on the same (or very similar) feature space; this MTL type is termed `homogeneous-feature MTL'. However, more elaborate schemes are also available for implementing the `heterogeneous-feature MTL' scenario \cite{zhang2017survey}\cite{kaiser2017one}, i.e. when the feature spaces (employed sets of sensors $S_{\beta}$) of the considered models are different and diverse in nature. The latter constitutes also a critical difference between the MTL and the transfer learning mechanisms.

\item {\it{Meta learning}}: The principal goal of the FL mechanism, as detailed in the beginning of Section \ref{ssec:FLLayer}, is to collectively create robust and powerful global AI models $\textbf{W}_\beta$ ((\ref{eq:FLGoblaModels})), by concatenating model updates $\Delta \textbf{W}_\beta^{l,m}$ ((\ref{eq:FLGoblaUpdate})) from multiple human teachers $h_m$ associated with the various network nodes $P_l$. However, there are application cases where the fundamental aim is not (only) to optimize the performance of the global models $\textbf{W}_\beta$, but rather to maximize the efficiency of the local ones $\textbf{W}_\beta^{l,m}$ ((\ref{eq:localModels})), e.g. when a new human teacher $h_m$ is introduced to the cognitive environment (and an accurate initialization of the respective $\textbf{W}_\beta^{l,m}$ models is required) or when the personalization ability of the overall system is critical (for example when developing customized human-robot interaction schemes). For addressing the latter requirements, the designed cognitive architecture incorporates means of so called meta learning techniques. Towards this direction, different meta learning methodologies have been proposed in the literature with Model Agnostic Meta Learning (MAML) receiving particular attention and demonstrating promising results in the deployment of FL-based systems \cite{jiang2019improving}\cite{fallah2020personalized}. According to the latter mechanism \cite{chen2018federated}, AI model parameter updates are estimated at two levels: a) inner update, where a set of human teachers $h_m$ (support set) are selected for computing a tentative updated version of $\textbf{W}_\beta$ and b) outer update, where the aforementioned tentative updated version of $\textbf{W}_\beta$ is evaluated over a separate set of human teachers $h_m$ (query set) for validating its efficiency and subsequently estimating an updated version $\textbf{W}'_\beta$ of the global models. In other words, the iteratively updated global models $\textbf{W}_\beta$ serve as an initialization step for constructing the local/personalized versions of $\textbf{W}_\beta^{l,m}$, which are computed using only the locally generated datasets $\delta^l$.
\end{itemize}

\subsection{Multi-agent human-robot collaborative learning}
\label{ssec:CollaborativeLearning}

Having introduced the constituent entities of the proposed cognitive AI architecture (namely the robotic platform, the human and the collective (FL) cognitive AI layer), their role and the corresponding formalisms, this sub-section summarizes how the ultimate goal of collaborative learning in a multi-agent human-robot environment is accomplished. In particular, knowledge/skill transfer (from the human side towards the robot one) is realized through the following two pathways: 
\begin{itemize}
\item {\it{Direct human-robot learning}}: Any human teacher $h_m$ can provide feedback to any employed robotic platform $P_l$ with respect to the functioning of AI model $\textbf{W}_\beta$, resulting into the creation of the local/customized model $\textbf{W}_\beta^{l,m}$, according to (\ref{eq:localModels}); the estimated $\textbf{W}_\beta^{l,m}$ can also be maintained and used as a `personalized' version of the global model $\textbf{W}_\beta$.
\item {\it{Direct robot-robot learning}} (or indirect propagation of collected feedback from all human teachers to all robotic platforms): Through the fundamental collective learning mechanism of FL (as explained in (\ref{eq:FLGoblaModels}) and (\ref{eq:FLGoblaUpdate})), knowledge ($\Delta \textbf{W}_\beta^{l,m}$) from all robotic platforms $P_l$ is combined (to produce an updated version $\textbf{W}'_\beta$ of each individual $ \textbf{W}_\beta$); the latter is subsequently shared/distributed back to all $P_l$ in the network.
\end{itemize}
Iterative execution of the above mechanism leads to the incarnation of a collaborative human-robot LfD learning environment/scheme.

Critical to the success of the overall LfD framework is the adoption of a teleoperation-based approach for collecting human feedback, making use of AR/VR visualization technologies (as detailed in Section \ref{ssec:Human}). Innovative aspect of the visualization interfaces in the proposed cognitive architecture, which differentiates them from similar ones of the literature, is that they are combined with XAI methods for explaining the decision-making process of AI models $\textbf{W}_\beta$. Specifically, the use of XAI techniques provides the privilege to the human teacher $h_m$ to be fully aware and in details of the exact factors that have led to a robotic malfunction (e.g. knowledge of the exact incorrectly classified pixels in a visual perception task). The latter inevitably enables the human teacher to provide more accurate and targeted guidance to the robot; hence, further boosting the capability of transferring fine-grained and sophisticated skills.

The learning capabilities of the proposed cognitive AI architecture are further reinforced by incorporating: a) means for modeling, interpreting and combining multi-user feedback information, namely user weighting, parameter weighting and user clustering (as described in Section \ref{sssec:UserIncorporation} and specified in (\ref{eq:FLGlobalUpdateHuman})-(\ref{eq:FLGlobalUpdateUserClustering})) and b) advanced AI learning techniques for estimating, modeling and exploiting cross-task correlation information, namely transfer, multi-task and meta learning (as detailed in Section \ref{sssec:TaskIncorporation} and explained in (\ref{eq:FLGlobalLossFL})-(\ref{eq:MultiTaskLearning})). It needs to be highlighted that the aforementioned approaches for integrating multi-user- and cross-task-related information can be flawlessly combined, leading to even more complex and sophisticated learning environments.

\section{CRM recovery case study}
\label{sec:CRMrecovery}

\begin{figure*}[th]
\centering
\includegraphics[width=1.00\textwidth]{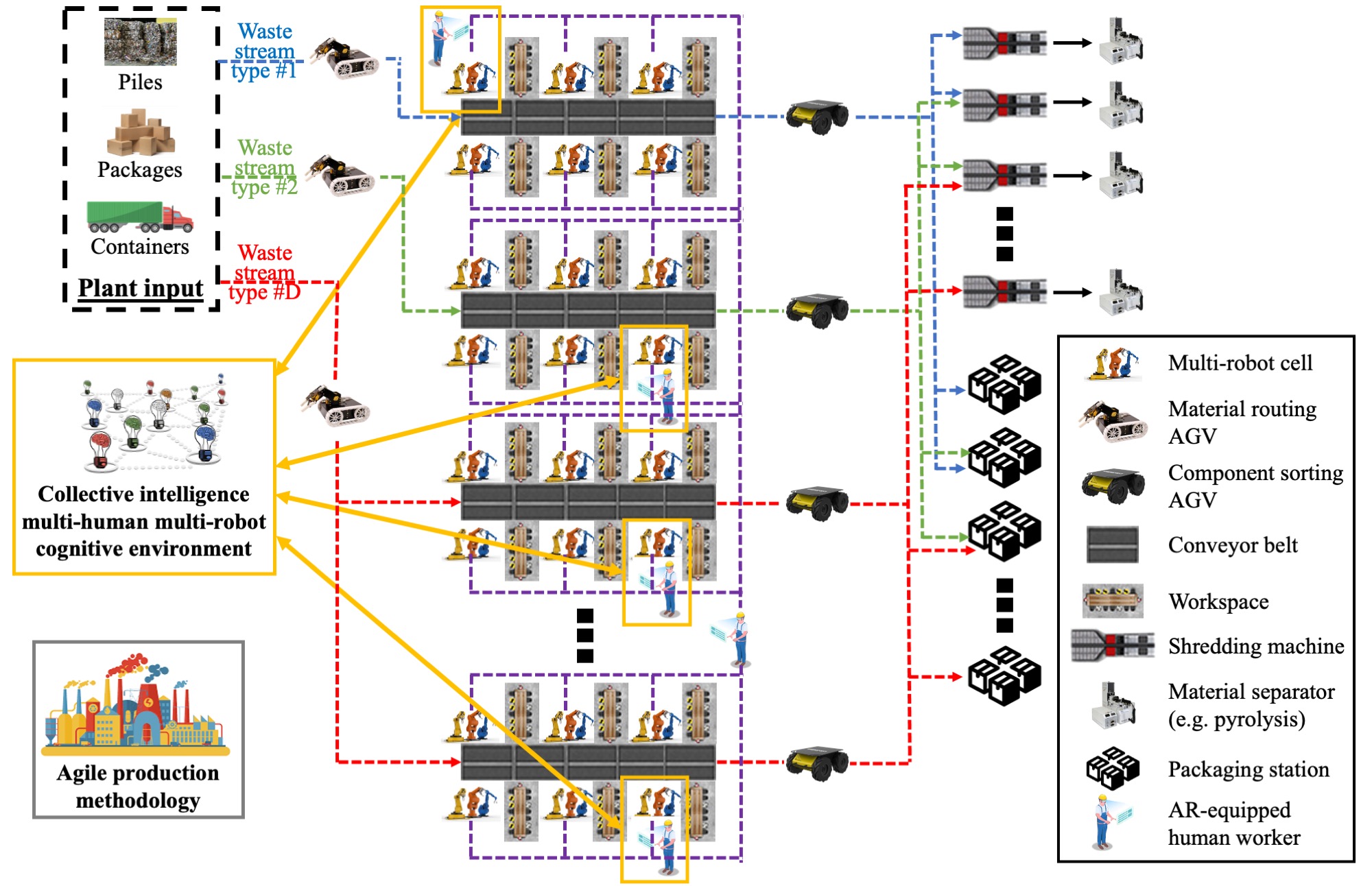}
\caption{Functional diagram of the designed cognitive AI architecture under an agile production CRM recovery case study.}
\label{f:CRMrecoveryUC}
\end{figure*}

In this section, the usefulness and benefits gained from the application of the designed cognitive AI architecture are demonstrated, by analysing an example of a real-world large-scale industrial manufacturing case study (subject to ongoing research activities) in the domain of Critical Raw Materials (CRM) recovery. Robotic platforms constitute a particularly suitable solution for the selected field, since they can undertake and automate numerous laborious, repetitive, tedious, stressful, harmful and hazardous human worker tasks, especially in the early stages of the recycling process (e.g. object dismantling, housing removal, component extraction, etc.). Overall, the CRM recovery pipeline inevitably needs to undergo an agile production operational methodology, since a) every recycling plant typically supports multiple waste streams of different type and nature, where the input materials arrive in an unsettled way and of significantly varying quantity, quality and composition, and b) the demands for the output recycled materials (posed by secondary market stakeholders, further recycling operators, etc.) also change (often rapidly) over time. The latter essentially poses the critical requirement for a high degree of adaptability to the involved robotic platforms, accompanied by the increased need for reinforced long-term autonomy (since new types of CRM materials and needed manipulations/operations are constantly encountered). To this end, the introduced cognitive AI architecture for multi-agent LfD learning, operating complementarily to a centralized factory-level orchestration methodology for agile production, constitutes an elegant choice for enabling the robotic platforms to continuously acquire new skills (from human demonstration). A high-level functional diagram of the envisaged agile production CRM recovery plant is illustrated in Fig. \ref{f:CRMrecoveryUC}.

\subsection{Robotized processing steps}
\label{ssec:RobotizedProcessingSteps}

Throughout the overall recycling plant operational pipeline, the input waste materials (e.g. laptops, personal computers, smartphones, tablets, TVs, batteries, etc.) undergo subsequent processing and manipulation steps, targeting to extract and group individual constituent device components (e.g. casings, plastics, capacitors, coolers, Printed Circuit Boards (PCBs), etc.) with homogeneous composition in terms of integrant critical raw materials (e.g. cobalt, lithium, phosphorus, magnesium, bauxite, etc.). For realizing the latter, different categories of robotic platforms with varying types of assigned tasks need to be deployed in each of the recycling plant's main processing steps (namely material routing, device dismantling and component sorting), as detailed in the followings.

\subsubsection{Material routing}
\label{sssec:MaterialsRouting}

The first step in the envisaged recycling plant work-cycle concerns how the different types of waste materials are being introduced to the processing pipeline. Under the current conceptualization, different means of plant input are foreseen (e.g. aggregated piles, received packages, delivered containers, etc.). The critical characteristic is that the input materials arrive in an unsettled way and in significantly varying pace, quantity, quality and composition. The latter inevitably poses significant challenges towards their efficient management, which in turn requires an overall continuous adjustment of the recycling plant work-plan and internal functionality. It needs to be highlighted that multiple types of waste streams (that contain valuable CRMs) are considered and processed in parallel, as mentioned in the beginning of Section \ref{ssec:RobotizedProcessingSteps}. Upon their receipt (and after unpacking, if needed), the different waste materials are identified/classified and assigned/introduced to a waste stream processing line. This step involves the transfer of the input waste materials to a dynamically assigned processing line of the factory. This dynamic `material routing' process involves the use of Automatic Guided Vehicles (AGVs), each mounted with a robotic arm (to implement the necessary pick and place actions) and a conveyor belt mechanism (for further facilitating the loading/unloading process). It needs to be highlighted that a varying number of processing lines for each individual waste stream type are also adaptively defined, according to the incoming waste status as well as the targeted overall plant output at each time instant.

\subsubsection{Device dismantling}
\label{sssec:DeviceDismantling}

Having performed the initial waste stream classification and its assignment to a processing line, the core building block of the overall solution is applied for realizing the `device disassembly/dismantling' step. In particular, each processing line comprises a conveyor belt, where the pace and speed of operation are dynamically controlled, taking into account the materials present or being processed. Along both sides of each conveyor belt, multiple workspaces are installed. The latter constitute the physical locations where the fine-grained manipulation of the input devices/materials (e.g. dismantling task) occurs. At each workspace, a multi-robot cell (consisting of multiple identical robotic arms) is dynamically assigned, based on run-time material recovery needs. The total number of multi-robot cells operating in each line is also adaptively and centrally defined, based on overall factory input/output targets and operational status. Each of the involved robot cells is equipped with a suitable tool changer, so as to support the use of multiple end-effectors and, hence, to enable different and diverse fine-grained manipulation tasks (e.g. cutting, unscrewing, drilling, breaking, etc.). The output of this step is a set of extracted and sorted constituent components of interest for every manipulated device, placed in appropriate collection baskets.

\subsubsection{Component sorting}
\label{sssec:ComponentSorting}
 
After the input waste materials are processed and the components of interest are extracted, the so called `component sorting' step is implemented, where AGVs are responsible for transferring the components from the disassembly line to the deposit location of the plant (e.g. packaging stations, material separation machines, etc.). The AGVs are equipped with effective basket mounting/unmounting and advanced SLAM navigation capabilities, in order to smoothly and safely operate in complex, time-varying environments and in the presence of humans.

\begin{figure*}[th]
\centering
\includegraphics[width=0.70\textwidth]{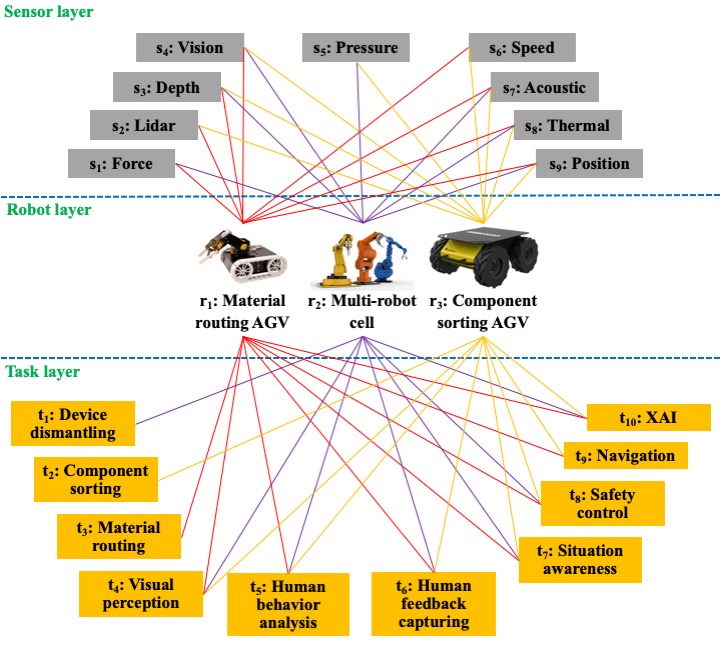}
\caption{Operational diagram of robotic platform types deployed under a CRM recovery case study.}
\label{f:CRMrecoveryUCRobots}
\end{figure*}

\subsection{Integration of the cognitive AI architecture}
\label{ssec:IntegrationCognitiveArchitecture}

In order to realize the above mentioned complex waste management activities in an autonomous, productive, efficient and safe way, each employed robotic platform $P_l$ is equipped with sophisticated edge computing AI-empowered cognitive technologies that reinforce its operational capabilities (e.g. working environment registration, human behavior analysis, waste manipulation, motion planning, navigation, human-robot interaction, safety control, etc.); the latter are implemented in the form of respective sophisticated AI models $\textbf{W}_\beta$. 

Following the formalization of the robotic platform $P_l$ ((\ref{eq:roboticPlatform})) in the introduced cognitive AI architecture (Section \ref{ssec:RoboticPlatform}), sets $S$, $R$ and $T$ are defined as follows to address the needs of the selected application case (resulting in a total of $3$ different types of platforms, i.e. an individual one for each robotized processing step): 
\begin{itemize}
\item $S=\{$\it{Force, Lidar, Depth, Vision, Pressure, Speed, Acoustic, Thermal, Position}$\}$
\item $R=\{$\it{Material routing AGV, Multi-robot cell, Component sorting AGV}$\}$
\item $T=\{$\it{Device dismantling, Component sorting, Material routing, Visual perception, Human behavior analysis, Human feedback capturing, Situation awareness, Safety control, Navigation, XAI}$\}$
\end{itemize}
The aforementioned elements, as well as their inter-connections towards forming the different types of robotic platforms $P_l$, are illustrated in the operational diagram in Fig. \ref{f:CRMrecoveryUCRobots}. It needs to be highlighted that sets $S$, $R$ and $T$ defined above include only a non-exhaustive collection of necessary elements for enabling the materialization of the envisaged recycling plant (i.e. additional types of sensors, robots and tasks can be incorporated for further improving and extending the targeted operational cycle), while multiple elements (e.g. motion planning ones, namely `Device dismantling', `Component sorting' and `Material routing') can be split into a series of more simple and specific ones (e.g. grasping, unscrewing, cutting, mounting, etc.). Moreover, it can be observed that the employed sensor, robot and task types are tightly interconnected, fact that motivates the extensive re-use of knowledge and where AI technologies (and in particular FL techniques) are especially suitable.

The most critical part for ensuring the long-term autonomy of the introduced system concerns the point where the robotic platforms $P_l$ and the human operators $h_m$ interact on the factory floor, targeting in principle the bootstrapping of the robotic activities as well as the acquisition of new skills from the side of the robotic agents. In particular, the human workers operate in a supervisory and proactive way, while they are equipped with AR technologies in order to receive real-time and accurate insights on the processes performed at the factory- and each individual workspace-level. The latter insights include detailed information on the status of the specific tasks performed by the robotic platforms (i.e. which steps have been implemented and which are planned to be performed) and regarding the robotic cognitive/inference operations (e.g. which objects have been recognized by the robot, how a given decision has been reached, what type of motion planning policies have been estimated, etc.). Especially for the latter case, the use of XAI tools enables the interpretation of the exhibited robot behavior, i.e. the identification of the root causes/procedures that led an AI-driven robotic platform to take specific decisions/actions. In this context, whenever a robot fails to complete a task, there is a high degree of uncertainty or the human worker identifies a deviation and decides to intervene, the robot pauses its operations and waits for human feedback. Subsequently, through the use of user-friendly and efficient AR technologies, the human worker is initially informed of the robot operational status (as described above), identifies the root cause of the possible malfunction and through the appropriate AR-based interaction/communication means guides the robot on how to successfully elaborate the task at hand. 

Apart from the in situ tuning phase, the principal goal of the overall system is for the robotic platforms to adaptively learn and in the long-term adapt their AI cognitive models $\textbf{W}_\beta$, based on the feedback received by the human workers (and modulated by the estimated user profiles $\textbf{Q}^{l,m}$). For incarnating this multi-human multi-robot collective intelligence vision, the introduced FL-based multi-agent LfD cognitive AI architecture (as detailed in Section \ref{sec:CognitiveArchitecture}) is applied. In particular, the proposed formalization will facilitate towards: a) incorporating multi-user feedback information for continuously updating the capabilities of the AI models $\textbf{W}_\beta$ (supporting user weighting, parameter weighting and user clustering techniques), as defined in (\ref{eq:FLGlobalUpdateHuman})-(\ref{eq:FLGlobalUpdateUserClustering}) and detailed in Section \ref{sssec:UserIncorporation} and b) integrating cross-task correlation information for further reinforcing the robustness of $\textbf{W}_\beta$ (relying on transfer, multi-task and meta learning methodologies), as formalized in (\ref{eq:FLGlobalLossFL})-(\ref{eq:MultiTaskLearning}) and explained in Section \ref{sssec:TaskIncorporation}. Moreover, it needs to be reminded that the designed FL-based scheme is by definition privacy-aware, since no data, apart from NN parameter updates, captured at each robot/node location $P_l$ are sent to the network.

\subsection{Insights and physical explanation of the introduced cognitive AI functions}
\label{ssec:InsightOnLfDFunctions}

In this section, detailed insights, physical interpretations and exemplary explanations (in the selected application domain) are provided, regarding the introduced AI-enabled cognitive functions. In particular, the proposed FL-based scheme for large-scale multi-agent human-robot collaborative learning (Section \ref{sec:CognitiveArchitecture}), which supports both direct human-robot and robot-robot knowledge transfer (Section \ref{ssec:CollaborativeLearning}), emphasizes on providing sophisticated mechanisms for incorporating a) multi-user feedback (Section \ref{sssec:UserIncorporation}) and b) cross-task correlation (Section \ref{sssec:TaskIncorporation}) information.

Regarding the combination of demonstration information from multiple users, the inherent (and often high) variance in the exhibited human behaviour (even for the same task) constitutes a critical challenge that is very likely to jeopardize the learning process of large-scale LfD systems. This difference in the observed human performance can be due to a number of factors, such as the variability in knowledge, idiosyncrasy, education, working experience, mental state, etc. across subjects. For efficiently overcoming the aforementioned obstacles, the introduced cognitive architecture comprises the following processes (that rely on the use of the $\textbf{Q}^{l,m}$ user profiles that model and interpret the behavior of each human teacher $h_m$): 
\begin{itemize}
\item {\it{User weighting}}: According to (\ref{eq:UserWeight}), the weight factor $\varepsilon (\textbf{Q}^{l,m})$ of each human teacher $h_m$ (at each robotic node $P_l$) is considered to be inversely proportional to the degree of dissimilarity of $\textbf{Q}^{l,m}$ with the respective global user model $\textbf{Q}^{g}$. The motivation behind the latter choice lies on the fundamental conceptualization that user profiles that deviate significantly from the global norm (i.e. being outliers) should receive decreased importance, for example, when specific user characteristics (e.g. working experience, educational background, etc.) need to receive particular significance/gradation, when significant inconsistencies are observed in the provided feedback instances (during the generation of $\textbf{W}_\beta^{l,m}$) of a given user (optionally also in relation to the remaining human teachers) or when the observed activity of an individual is frequently detected to be related with undesirable user states (e.g. when in extreme situations of fatigue, stress, decreased attention, etc.).
\item {\it{Parameter weighting}}: The formalism in (\ref{eq:FLGlobalUpdateParametersWeighting}) modulates the knowledge aggregation step, taking into account the cross-correlations among the parameters of the user profile $\textbf{Q}^{l,m}$ and the AI $\textbf{W}_\beta^{l,m}$ models. Specifically, when such strong correlations are identified (i.e. increased $\textbf{r}^{l,m}_\beta$ values), these are emphasized when creating the updated version of the global AI model $\textbf{W}'_\beta$. The physical explanation of the latter implies that when aspects of the exhibited behavior of a given teacher $h_m$ are detected to be consistently related with parts of a certain AI model $\textbf{W}_\beta$ (e.g. when a specific individual is identified to provide credible feedback for the detection of (parts of) specific object classes in a visual perception task or the execution of (parts of) particular policies in a motion planning scenario), this should be considered as reliable/critical information that must subsequently be appropriately amplified during the knowledge aggregation step. Such a mechanism is particularly beneficial for prioritizing the collected feedback information at a finer level of detail and addressing misalignments/inconsistencies/conflicts among different users.
\item {\it{User clustering}}: The mechanism defined in (\ref{eq:FLGlobalUpdateUserClustering}) aims at tackling the case of high heterogeneity in the provided feedback information, i.e. when the conventional (in FL-based systems) Independent and Identically Distributed (IID) data assumption is considered very likely to be violated. In particular, the fundamental consideration of this mechanism states that there does not exist a single global model $\textbf{W}_\beta$ for a given robotic task that can adequately capture the shared knowledge of all human teachers $h_m$. The latter is particularly applicable for tasks where essentially significantly diverse user behaviors can be observed (e.g. when specifying the movement trajectory of a mobile robot on the factory floor, defining the robot safety control policy that also reckons (future) human intentions, etc.). For accounting for these aspects, multiple instances $\textbf{W}_{\beta,\eta}$ of each global model $\textbf{W}_\beta$ are created and, apart from modelling robust $\textbf{W}_{\beta,\eta}$ models, an optimal matching among human teachers $h_m$ (at each local network node $P_l$) and specific $\textbf{W}_{\beta,\eta}$ instances is simultaneously learned (indicated by factors $\theta^{l,m}_{\beta,\eta}\in\{0,1\}$). Eventually, feedback information provided by human teacher $h_m$ is used to only update the associated $\textbf{W}_{\beta,\eta}$ instance.
\end{itemize}
It needs to be reminded that the above multi-user feedback fusion schemes can also be flawlessly combined in a single FL-based LfD system. 

With respect to the modelling and exploitation of cross-correlations among different robotic tasks, this has to be inevitably addressed in real-world application scenarios (where multiple tasks are typically co-occurring), while even relatively small improvement in performance (of each individual task) can have a potentially large operational/economic impact, especially in large-scale industrial applications. For addressing the latter, the developed cognitive architecture incorporates the following advanced AI-empowered learning methods:
\begin{itemize}
\item {\it{Transfer learning}}: The formalism defined in (\ref{eq:FLGlobalLossFL}) and (\ref{eq:FLGlobalLossFLTransfer}) aims at estimating the inter-dependencies and the joint modelling of pairs of robotic tasks that share similarities in the underlying feature space (i.e. operate using the same or similar types of sensorial data). In particular, the overall goal is to realize knowledge re-use during the learning process, i.e. to transfer, combine and integrate learned knowledge structures that have been formulated for model $\textbf{W}_{\beta 1}$ to improve the performance of an operationally similar model $\textbf{W}_{\beta 2}$ (and vice versa). The latter characteristic is particularly useful when sufficiently large training datasets are available for model $\textbf{W}_{\beta 1}$ (and the respective learning process can be completed successfully), but not for the functionally similar model $\textbf{W}_{\beta 2}$ (whose training procedure can be reinforced by transferring knowledge gained when constructing $\textbf{W}_{\beta 1}$). For example, in the selected application domain, the goals of the visual perception models associated with the `Material routing AGV'  and the `Multi-robot cell' (Fig. \ref{f:CRMrecoveryUCRobots}) robotic platforms present significant similarities; the former targets to detect whole device types in the early stages of the `Material routing' process, while the latter has as a principal goal to identify the individual constituent components of each such device during the `Device dismantling' procedure. Another similar example concerns the `Device dismantling' and `Component sorting' robotic tasks (Fig. \ref{f:CRMrecoveryUCRobots}), where, despite the different end goals and particularities of each process, the underlying robot motion planning policies inevitably share similar patterns and primitive actions that the joint learning of the respective AI models would be beneficial.
\item {\it{Multi-task learning}}: The mechanism detailed in (\ref{eq:MultiTaskLearningLoss}) and (\ref{eq:MultiTaskLearning}) shares similarities with the `transfer learning' one described above, but exhibits the following key differences: a) It aims to simultaneously construct multiple reliable $\textbf{W}_\beta$ models (by exploiting their cross-correlations), rather than only transferring knowledge cues between different models, b) It can involve $\textbf{W}_\beta$ models that operate on the same (or very similar) feature space (`homogeneous-feature MTL'), but also on different/diverse ones (`heterogeneous-feature MTL' ), c) It allows, apart from the explicit a priori definition of the AI model pairs with similar operations, also the automatic learning of the task correlation weights (i.e. precision matrix $\mathbf{\Omega}$) directly from the data, while it has been experimentally shown to perform well also when considering uncorrelated tasks. Within the context of CRM recovery, for example, the `homogeneous-feature MTL' mechanism can be adopted to jointly construct the visual perception (or the navigation) $\textbf{W}_\beta$ models of the `Material routing AGV'  and the `Component sorting AGV' (Fig. \ref{f:CRMrecoveryUCRobots}), which operate at the beginning and the end of the overall factory process pipeline, respectively; however, presenting similar operational characteristics. On the other hand, the `heterogeneous-feature MTL' scheme can simultaneously build the visual perception and the device dismantling $\textbf{W}_\beta$ models of the `Multi-robot cell' (or the visual perception and the material routing ones of the `Material routing AGV'); the latter correspond to co-occuring robotic tasks (using diverse types of sensorial data), yet exhibiting significant inter-dependencies/relations during their execution. 
\item {\it{Meta learning}}: The fundamental conceptualization of this scheme is to optimize the performance of the local AI models $\textbf{W}_\beta^{l,m}$, rather than (only) to maximize the efficiency of the global $\textbf{W}_\beta$. Such a routine is beneficial, for example, when a new human worker $h_m$ enters the factory floor in the selected application case (and the corresponding $\textbf{W}_\beta^{l,m}$ models need to be accurately initialized) or when the performance of the personalized $\textbf{W}_\beta^{l,m}$ models is of paramount importance (e.g. realizing personalized safety control at all stages of the overall CRM recovery process).
\end{itemize} 

\section{Conclusions and future research directions}
\label{sec:Conclusions}

In this paper, the problem of robotic Learning from Demonstration (LfD) was thoroughly investigated and a novel cognitive architecture for large-scale robotic learning was introduced for enabling the robust deployment of open, scalable and expandable robotic systems in large-scale and complex environments. The fundamental consideration of the designed architecture is grounded on the establishment of a Federated Learning (FL)-based framework for implementing a multi-human multi-robot collaborative learning environment. Regarding pivotal novelties, the designed cognitive architecture (that significantly broadens the capabilities of current LfD robotic learning schemes): a) introduces a new FL-based formalism that extends the conventional LfD learning paradigm to support large-scale multi-agent operational settings, b) elaborates previous FL-based self-learning robotic schemes so as to incorporate the human in the learning loop, and c) consolidates the fundamental principles of FL with additional sophisticated AI-enabled learning methodologies for modelling the multi-level inter-dependencies among the robotic tasks. The applicability of the designed framework was explained through an example of a real-world industrial case study for agile production-based Critical Raw Materials (CRM) recovery. 

In the followings, the key elements of the designed environment that are crucial to its success, which at the same time constitute both critical technological challenges and promising future research directions in the fields of LfD, AI and HRI, are briefly summarized:

\begin{itemize}

\item {\it{Incorporation of multi-user feedback information}}: Although numerous LfD approaches have been introduced in the past, multi-agent learning remains as one of the most challenging open issues in the field \cite{Hussein17}. The latter comprises the central goal of the introduced cognitive AI architecture, which relies on the incorporation of FL-based technologies reinforced with a generalized function $\Phi(.)$ (defined in (\ref{eq:FLGlobalUpdateHuman})) for specifying how to combine information (i.e. model updates $\Delta \textbf{W}_\beta^{l,m}$) from multiple human subjects $h_m$ (making use of the respective estimated profiles $\textbf{Q}^{l,m}$). Three variants of function $\Phi(.)$ are introduced (Section \ref{sssec:UserIncorporation}), namely user weighting, parameter weighting and user clustering. The specific choice to be made depends on the particularities of the selected application domain (e.g. the number of involved human workers, the number and types of supported AI models, the type and number of employed sensors, etc.).

\item {\it{Incorporation of cross-task correlation information}}: LfD methods have so far focused in principle on introducing invariance to the observed human behavior for a given task. However, the designed architecture puts emphasis on exploiting the correlations among different tasks (i.e. among the various $\textbf{W}_\beta$ or $\textbf{W}_\beta^{l,m}$ models); hence, significantly elaborating current LfD principles. Three individual methodologies are proposed towards this direction (Section \ref{sssec:TaskIncorporation}), namely transfer, multi-task and meta learning. The selection of the specific approach to be used (or a combination of them), depends again on the individual characteristics and goals of the targeted application (e.g. number and type of $\textbf{W}_\beta$ models involved, if new human users are expected to be involved and how often, etc.). 

\item {\it{DL-empowered user profiling}}: There has been an extensive body of research activity for creating robust user profiles, using various ML techniques, over the recent years. However, further focusing on implementing such models following the DL paradigm would likely lead to significant performance gains and increased robustness, while, importantly, the use of NNs \cite{farnadi2018user}\cite{liang2020drprofiling} for building $\textbf{Q}^{l,m}$ models ((\ref{eq:FLGlobalUpdateHuman})) would enable the incarnation of fully end-to-end trainable systems and $\textbf{Q}^{l,m}$ to be constructed through the designed FL-based cognitive environment. Further challenges comprise the use of multi-modal information from multiple non-invasive sensorial devices, while simultaneously modelling the variance in the exhibited human behavior (among the same or different individuals).

\item {\it{Deployment of FL technologies in robotic systems}}: FL itself often poses significant deployment challenges of various types (e.g. low connectivity, increased number of network nodes, latency in communication, etc.) \cite{konevcny2016federated}\cite{Bonawitz19}\cite{yan2020risk}. In the context of a large-scale industrial robotic setting, such challenges will have an increased importance, while the expected levels of system robustness and interoperability will inevitably need to meet high industrial standards as well. To this end, reliable FL-based solutions would likely need to capitalize on, apart from algorithmic optimizations of the FL mechanism, the capabilities provided by additional emerging technologies (e.g. 5G/6G network connectivity, quantum computing, hardware AI implementations, etc.).

\item {\it{Human-centred eXplainable AI}}: Over the recent years, an extensive body of research has been devoted on investigating various XAI methodologies, resulting in numerous diverse approaches and promising results \cite{arrieta2020explainable}. However, the focus has so far been placed on addressing the `explainability' aspect (i.e. to identify the underlying reasons for the exhibited behavior of the AI models), leaving the `interpretability' perspective (i.e. the human users actually understanding the observed AI behavior) largely under-explored \cite{adadi2018peeking}. The latter becomes even more demanding for cases of non-IT human experts being involved. A promising direction would naturally require the incorporation of Human-Computer Interaction (HCI) and human sciences principles, e.g. through the use of dynamic visualizations, question-answering schemes, interactive mechanisms, etc.

\item {\it{Addressing of (cyber-)security, personal data and privacy/ethics issues}}: Despite the fact that FL is a by-definition privacy aware approach that requires no exchange of actual data (only AI model parameter updates are transmitted) among the network nodes, significant research efforts have been devoted recently towards addressing possible security and privacy preserving gaps (e.g. differential \cite{Geyer17}, model-poisoning \cite{bagdasaryan2020backdoor}, white-box inference \cite{Nasr19} attacks, etc.). The latter are often combined with innovative techniques or emerging technologies, such as differential privacy \cite{kim2020incentive}, homomorphic encryption \cite{song2020privacy}, blockchain \cite{rahman2020secure}, etc. However, explicitly modeling and integrating human user feedback information, through the creation of user profiles $\textbf{Q}^{l,m}$ and their incorporation in the FL mechanism (Section \ref{ssec:FLLayer}), inevitably requires the elaboration and extension of the aforementioned methodologies.
\end{itemize}

\section*{Acknowledgments}
The work presented in this paper was supported by the ICS-FORTH internal RTD Programme `Ambient Intelligence and Smart Environments'.

\balance
\bibliographystyle{IEEEtran}
\bibliography{arXiv_v1}

\end{document}